\documentclass[preprint,12pt,authoryear]{elsarticle}

\usepackage{amssymb}
\usepackage{adjustbox}
\usepackage{graphicx}
\usepackage{multirow}
\usepackage{booktabs}
\usepackage{ifthen}
\usepackage{subcaption}
\usepackage{tikz}
\usepackage{xcolor}

\emergencystretch=\maxdimen
\usepackage{lineno}

\makeatletter
\def\ps@pprintTitle{%
   \let\@oddhead\@empty
   \let\@evenhead\@empty
   \def\@oddfoot{\reset@font\hfil\thepage\hfil}
   \let\@evenfoot\@oddfoot
}
\makeatother

\journal{ISPRS Journal of Photogrammetry and Remote Sensing Articles}

\biboptions{authoryear}

\begin{document}
\begin{frontmatter}



\title{Multi-task fully convolutional network for tree species mapping in dense forests using small training hyperspectral data}




\author[1,2]{Laura Elena Cué La Rosa\corref{cor1}}\ead{lauracue@aluno.puc-rio.br}
\author[3]{Camile Sothe}\ead{sothec@mcmaster.ca}
\author[1]{Raul Queiroz Feitosa}\ead{raul@ele.puc-rio.br}
\author[4]{Cláudia Maria de Almeida}\ead{almeida@dsr.inpe.br}
\author[5]{Marcos Benedito Schimalski}\ead{marcos.schimalski@udesc.br}
\author[6]{Dario Augusto Borges Oliveira}\ead{dariobo@br.ibm.com}

\cortext[cor1]{Corresponding author}
\address[1]{Computer Vision Lab, Pontifical Catholic University of Rio de Janeiro, Rio de Janeiro, Brazil}
\address[2]{IBM Research, Avenida Pasteur, 138, Rio de Janeiro, Brazil}
\address[3]{School of Earth, Environment {\&} Society, McMaster University, 1280 Main Street West, Hamilton, Ontario, Canada}
\address[4]{Department of Remote Sensing, National Institute for Space Research (INPE), Av. dos Astronautas 1758, São José dos Campos SP, Brazil}
\address[5]{Department of Forest Engineering, College of Agriculture and Veterinary, Santa Catarina State University (UDESC), Avenida Luiz de Camões 2090, Lages, SC, Brazil}
\address[6]{IBM Research, Rua Tutóia, 1157, São Paulo, Brazil}

\begin{abstract}

This work proposes a multi-task fully convolutional architecture for tree species mapping in dense forests from sparse and scarce polygon-level annotations using hyperspectral UAV-borne data. Our model implements a partial loss function that enables dense tree semantic labeling outcomes from non-dense training samples, and a distance regression complementary task that enforces tree crown boundary constraints and substantially improves the model performance. Our multi-task architecture uses a shared backbone network that learns common representations for both tasks and two task-specific decoders, one for the semantic segmentation output and one for the distance map regression. We report that introducing the complementary task boosts the semantic segmentation performance compared to the single-task counterpart in up to 11\% reaching an average user's accuracy of 88.63\% and an average producer's accuracy of 88.59\%, achieving state-of-art performance for tree species classification in tropical forests.

\end{abstract}

\begin{keyword}
Semantic segmentation \sep 
Tree species identification \sep 
Multi-task learning \sep
Fully convolutional network \sep 
Sparse annotations
\end{keyword}

\end{frontmatter}

\section{Introduction}\label{intr}
Mapping tree species using remote sensing has consolidated as a cheaper, faster, and more practical way to inventory forest areas in comparison with traditional fieldworks \citep{Fassnacht2016}. Identifying individual trees even in dense forest canopies of tropical forests is currently possible due to improved spatial and spectral resolution of remote sensing data associated with the increase in computational capacity and the advancements in classification methods. Hyperspectral data collected by airborne platforms or unmanned aerial vehicles (UAV) enable the discrimination of individual tree crowns (ITC) and the classification of tree species in tropical environments by capturing slight differences in reflectance patterns among them \citep{clark2012species,feret2012tree,baldeck2015operational,ferreira2016mapping,shen2017tree,sothe2019tree,zhang2020three}.

When using hyperspectral or multisource data, most studies involving tree species classification apply machine learning algorithms, such as support vector machine (SVM) and random forest (RF) \citep{Fassnacht2016}. Such algorithms are robust and usually work well for different class distributions or high dimensional data \citep{ghosh2014framework}. However, most of them depend on intricate heuristics that limit their transferability, and often it is hard to achieve an optimal balance between discrimination and robustness for many types of data \citep{zhang2016deep,zhang2020three}. 

Deep learning methods proved to be a robust alternative for remote sensing image classification, as they can learn optimal features and classification parameters to handle hyperspectral data \citep{signoroni2019deep}. Different works successfully applied convolutional neural networks (CNN) \citep{NIPS2012_c399862d} for tree species classification \citep{polonen2018tree,hartling2019urban,fricker2019convolutional,natesan2020individual,mayra2021tree}, including tropical and subtropical environments \citep{sothe2019tree,sothe2020comparative,zhang2020three,tian2020improved,abbas2021characterizing}. \cite{sothe2019tree} employed a CNN based on image patches classification to classify hyperspectral data pixels and reached significantly higher accuracies (84.4\%) when compared to SVM (62.7\%) and RF (59.2\%). \citet{zhang2020three} proposed a three-dimensional CNN (3D-CNN) for tree species classification in hyperspectral data, reaching an accuracy of 93.14\%. \citet{mayra2021tree} also used a 3D-CNN in the field of hyperspectral image analysis to classify three major tree species and a keystone species, European aspen, characterized by a sparse and scattered occurrence in boreal forests. Recently, \citet{abbas2021characterizing} explored the use of a CNN to discriminate among 19 urban tree species from hyperspectral data, achieving an overall accuracy higher than 85\%. However, these approaches are inefficient for large-scale remote sensing imagery, as they infer each pixel classification using its corresponding patch (surrounding neighboring pixels). 

More efficient approaches use Fully Convolutional Networks (FCN) \citep{long2015fully}, which classify all pixels in the input patch at once. Currently, state-of-the-art methods for semantic segmentation in remote sensing images are based on FCN architectures \citep{zhu2017deep,ma2019deep}, including tree species identification and segmentation from UAV-RGB images \citep{kattenborn2019convolutional,lobo2020applying,ferreira2020individual,brandt2020unexpectedly,schiefer2020mapping}, but only few of them used hyperspectral data. In \citet{wagner2019using}, the authors assess a U‐net FCN to identify and segment one tree species, \textit{Cecropia hololeuca}, using very high-resolution images, obtaining an overall accuracy of 97\% and an intersection over union (IoU) of 0.86. \cite{fricker2019convolutional} employed FCN to classify tree species in a mixed-conifer forest from hyperspectral and pseudo-RGB data reporting an average F-score of 0.87 and 0.64, respectively. \cite{lobo2020applying} evaluated the performance of several variants of FCNs combined with a conditional random field (CRF) post-processing step for single tree species in an urban environment from high-resolution UAV optical imagery, reporting an overall accuracy ranging from 88.9\% to 96.7\%. In \citet{ferreira2020individual}, the authors employed an encoder composed of several residual blocks \citep{he2016deep} and a decoder based on an Atrous Spatial Pyramid Pooling (ASPP) \citep{chen2017deeplab} for individual tree detection and tree species classification of Amazonian palms in UAV-RGB images. The authors further proposed a post-processing step based on morphological operations for boundary refinement and individual trees separation. Considering hyperspectral images, \citet{miyoshi2020novel} proposed a novel FCN method to identify single-tree species in highly-dense forest of the Brazilian Atlantic biome. Different from other works, that deliver a semantic labeled image, \citet{miyoshi2020novel} proposed a FCN that produce a confidence map of the trees location. In \citet{ferreira2021accurate}, the authors tested three backbones networks for feature extraction incorporated into a DeepLabv3+ decoder \citep{chen2018encoder} using WorldView-3 satellite images to map Brazil nut trees, reporting a producer's accuracy higher than 93\%. In addition the author tested the model robustness reducing the percentage of training images patches. Recently, \citet{hao2021automated} successfully applied a mask region-based convolutional neural network (Mask R-CNN) for detecting Chinese fir’s individual tree crown and height. However, as pointed out by the authors, this method struggle in scenarios with highly overlapping crowns.

The studies mentioned above depends on a large number of densely annotated ITC training samples (i.e, all pixel's labels within the input image are known) to deliver an accurate semantic segmentation map, which may restrict their application in classifying a large number of tree species in diverse dense forests. For instance, in tropical forests, tree species are not equally distributed over the forest canopy where, for a given region, some might be dominant and others rare. That often results in an imbalanced training set, where only a few samples are available for under-represented classes \citep{mellor2015exploring,Fassnacht2016}. Besides, commonly, the few ITC samples are sparsely distributed across the images \citep{Fassnacht2016}, which represents an additional challenge for such methods. Sampling considering the natural abundance of species leads to severely imbalanced training sets, but increasing the sample size of rare species is, conversely, time-consuming and costly \citep{graves2016tree}. Thus, a reasonable number of samples per species required by many methods to perform optimal classification in tropical forests is rarely reached \citep{feret2012tree}.

In the last few years, some alternatives arose to train FCN with weak supervision, enabling low-cost annotations using points and scribbles \citep{alonso2017coral,maggiolo2018improving,wu2018scribble,tang2018normalized}. In \citet{tang2018normalized}, the authors proposed to train a FCN from scribbles using a partial cross-entropy (pCE) loss. The proposed loss only back-propagates gradients for the scribble annotated pixels, emerging as an effective approach to deal with low-cost annotations. Considering remote sensing applications, \cite{wu2018scribble} employed the same loss function to train an FCN for segmenting aerial building footprints, achieving an IoU of 68.4\% with only 5\% scribble samples. The mentioned weakly supervised learning methods are yet to be explored for tree species semantic segmentation in dense forests, considering that collecting samples in this application is particularly costly, time-consuming, and requires specific domain expertise. 

In a previous work \citep{sothe2020comparative}, we compared conventional CNN, SVM, and RF methods and different classification approaches (pixel, object, and majority-vote rule), reporting the CNN patch-based model as the overall most accurate. This work builds on that and proposes a novel network architecture to train an FCN with sparse and scarce polygon-level annotations (ITC samples) for tree species mapping in dense forest canopies using the hyperspectral UAV-borne data explored in \citet{sothe2020comparative}. Our FCN method is less computationally demanding and, at the same time, delivers similar or better results than the previously proposed CNN approach. First, we propose implementing a partial loss function to train FCN to perform dense tree semantic labeling from non-dense training samples. Second, we modify the architecture to implement a distance regression complementary task that substantially improves the model performance by enforcing tree crown boundary constraints. The proposed multi-task fully convolutional architecture uses a shared backbone network that learns common representations for both tasks and two task-specific decoders, one for the semantic segmentation output and one for the distance map regression. We report that introducing the complementary task boosts the semantic segmentation performance compared to the single-task counterpart. Our Multi-Task Fully Convolutional Network for Sparse Polygon-level annotations (MTFCsp) network is trained end-to-end and delivers dense predictions. 

The paper is organized as the following. Section \ref{dataset} describes the study area. Section \ref{method} details the proposed method. Section \ref{exp} presents the experimental design and evaluation measures. Sections \ref{results} and \ref{disc} present the results and discussion, respectively. Finally, Section \ref{conclusion} presents the main conclusions of this paper.

\section{Materials}\label{dataset}

\subsection{Study Area}
\begin{figure*}
	\centering
		\includegraphics[width=1.0\linewidth]{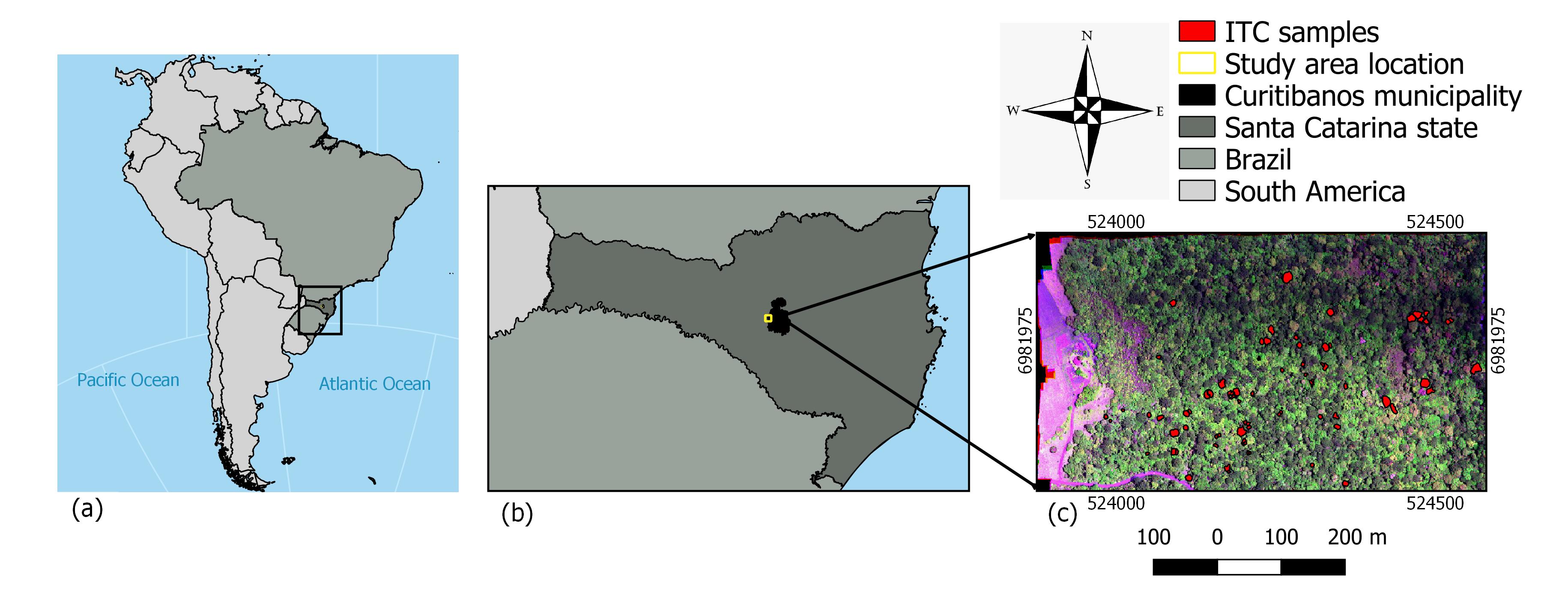}
	\caption{Study area location. (a) Brazil and South America, (b) Santa Catarina state and Curitibanos municipality, (c) Hyperspectral image with the location of individual tree crown (ITC) samples.}
	\label{fig:data}
\end{figure*}

The study area is located in the municipality of Curitibanos, Santa Catarina state, a southern region of Brazil (Figure \ref{fig:data}). The area comprises an extension of approximately 30 ha and belongs to the Atlantic Rain Forest biome and the Mixed Ombrophilous Forest phytophysiognomy. The region is dominated mainly by broadleaves species, having only two conifer species. One of them, \textit{Araucaria angustifolia}, is a physiognomic marker of this forest type \citep{backes1983araucaria}.

\subsection{Hyperspectral images}
The hyperspectral images were acquired from a frame format camera based on a Fabry-Perot Interferometer (FPI), model 2015 (DT-0011) boarding a quadcopter UAV (UX4 model). The camera has two CMOSIS CMV400 sensors with an adjustable air gap that can be flexibly configured to select up to 25 spectral bands between 500 and 900 nm with the minimum bandwidth of 10 nm at the full width at half maximum (FWHM) \citep{kwok2018ecology} (see configuration in Table \ref{tab:bandwith}). The first preprocessing stage of the images comprised the radiometric calibration, in which digital numbers were transformed into radiance values, and the dark signal correction. This step was performed in the software Hyperspectral Imager provided by \citet{senop2017datasheet} using a black image collected with the lens covered right before the data capture \citep{de2016geometric}. The geometric processing involved each band's orientation using the interior orientation parameters (IOPs) and the exterior orientation parameters (EOPs), which were estimated using the so-called on-the-job calibration. For the camera positions, the initial values were assessed by the GNSS receiver and involved latitude, longitude, and altitude (flight height plus the average terrain elevation) data. The coordinates of six ground control points (GCPs) were added to the project and measured in the corresponding reference images in the sequence. These points were previously located and surveyed in the field (signalized with lime mortar) on the same day of the flight and had their coordinates collected with a GNSS RTK Leica GS15. After the bundle adjustment, the final errors in the GCPs (reprojection errors) were 0.03 pixels in the image and 0.003 m in the GCPs. Finally, the orthorectification of each band was performed separately. This process also co-registers the bands of the same image regarding their slight positioning difference caused by the camera's time-sequential operating principle \citep{honkavaara2013processing}. The final dataset is composed of 25 spectral bands with 11 cm of spatial resolution. More details about the flight, camera, and preprocessing steps can be found in \cite{sothe2020comparative}.

\begin{table}
    \caption{Spectral settings for the hyperspectral camera ($\lambda$ = central wavelength of the spectral band).}\label{tab:bandwith}
    \begin{adjustbox}{width=0.98\textwidth,center}
    \begin{tabular}{cc|cc|cc|cc|cc}
    \toprule
    $\lambda$ & FWHM & $\lambda$ & FWHM & $\lambda$ & FWHM & $\lambda$ & FWHM & $\lambda$ & FWHM\\
    (nm) & (nm) & (nm) & (nm) & (nm) & (nm) & (nm) & (nm) & (nm) & (nm)\\
    \midrule
    506 & 15.65	& 580 & 15.14 & 650 & 15.85 & 700 & 21.89 & 750 & 19.43\\
    519 & 17.51 & 591 & 14.81 & 659 & 24.11 & 710 & 20.78 & 769 & 19.39\\
    535 & 16.41 & 609 & 13.77 & 669 & 21.7 & 720 & 20.76 & 780 & 18.25\\
    550 & 15.18 & 620 & 14.59 & 679 & 21 & 729 & 21.44 & 790 & 18.5\\
    565 & 16.6 & 628 & 12.84 & 690 & 21.67 & 740 & 20.64 & 819 & 18.17\\
    \bottomrule
    \end{tabular}
    \end{adjustbox}
\end{table}

\subsection{Individual tree crown samples}
Samples of 14 species totaling 70 ITCs were acquired in fieldwork conducted in December 2017 (Table \ref{tab:tree}). Only ITCs visited and identified in the fieldwork were used as samples for semantic segmentation. Overlapping ITCs or trees with ambiguous appearance were discarded. In addition, we used images collected by an UAV-RGB camera with 4 cm spatial resolution to aid the delineation of the ITC samples and to avoid including pixels from surrounding trees. Due to the dense forest canopy with many species in a small area, it was not possible to collect a large number of samples for each species. This resulted in an imbalanced sample set sparsely distributed over the study area, with dominant species having 8 to 9 ITC samples and the minority ones having only 2 ITCs. The available ITCs samples  were divided into 54\% training and 46\% test sets, always keeping the ITC identity. The number of ITC and pixel samples for each set is depicted in Table \ref{tab:tree}.

\begin{table}[!ht]
    \caption{List of tree species: species names, number of ITCs and number of pixels for training and test sets.}\label{tab:tree}
    \begin{adjustbox}{width=0.95\textwidth,center}
    \begin{tabular}{llllllll}
    \toprule
    & & \multicolumn{2}{c}{Total} & \multicolumn{2}{c}{Training} & \multicolumn{2}{c}{Test}\\
    ID & Species names & ITCs & pixels & ITCs & pixels & ITCs & pixels\\
    \midrule
    a &\textit{Luehea}           & 3     & 23,624    & 2 & 8,969       & 1  & 7,582\\
    b &\textit{Araucaria}        & 8     & 27,191    & 4  & 13,330     & 4  & 13,928\\
    c &\textit{Mimosa}           & 6     & 25,449    & 3  & 9,362      & 3  & 9,901\\
    d &\textit{Lithraea}         & 5     & 17,458    & 3  & 7,185      & 2  & 10,290\\
    e &\textit{Campomanesia}     & 5     & 18,837    & 3  & 12,070     & 2  & 6,753\\
    f &\textit{Cedrela}          & 5     & 24,368    & 3  & 12,029     & 2  & 12,333\\
    g &\textit{Cinnamodendron}   & 2     & 6,927     & 1  & 2,294      & 1  & 1,321\\
    h &\textit{Cupania}          & 2     & 12,475    & 1  & 5,147      & 1  & 2,113\\
    i &\textit{Matayba}          & 7     & 48,231    & 4  & 25,466     & 3  & 20,290\\
    j &\textit{Nectandra}        & 8     & 11,247    & 4  & 4,805      & 4  & 6,539\\
    k &\textit{Ocotea}           & 9     & 101,884   & 5  & 62,928     & 4  & 38,939\\
    l &\textit{Podocarpus}       & 6     & 12,387    & 3  & 6,513      & 3  & 5,899\\
    m &\textit{Schinus sp1}      & 2     & 4,491     & 1  & 2,142      & 1  & 2,341\\
    n &\textit{Schinus sp2}      & 2     & 6,083     & 1  & 5,096      & 1  & 1,001\\\midrule
    Total               &        & 70    & 340,652   & 38 & 194205     & 32 & 146447\\
    \bottomrule
    \end{tabular}
    \end{adjustbox}
\end{table}

\begin{figure}[htbp]
\begin{subfigure}{.5\textwidth}
  \centering
  \includegraphics[width=1\linewidth]{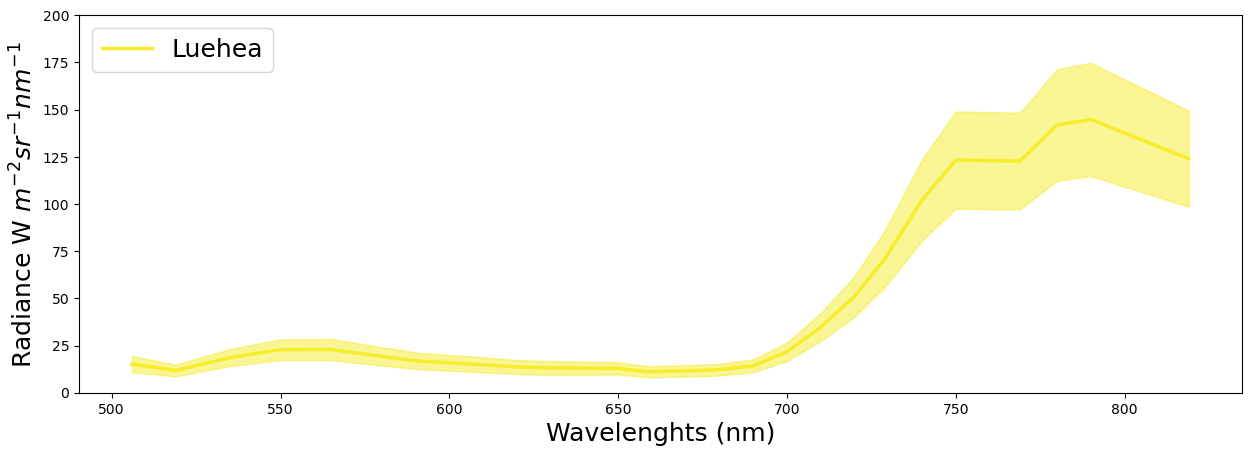}
\end{subfigure}%
\begin{subfigure}{.5\textwidth}
  \centering
  \includegraphics[width=1\linewidth]{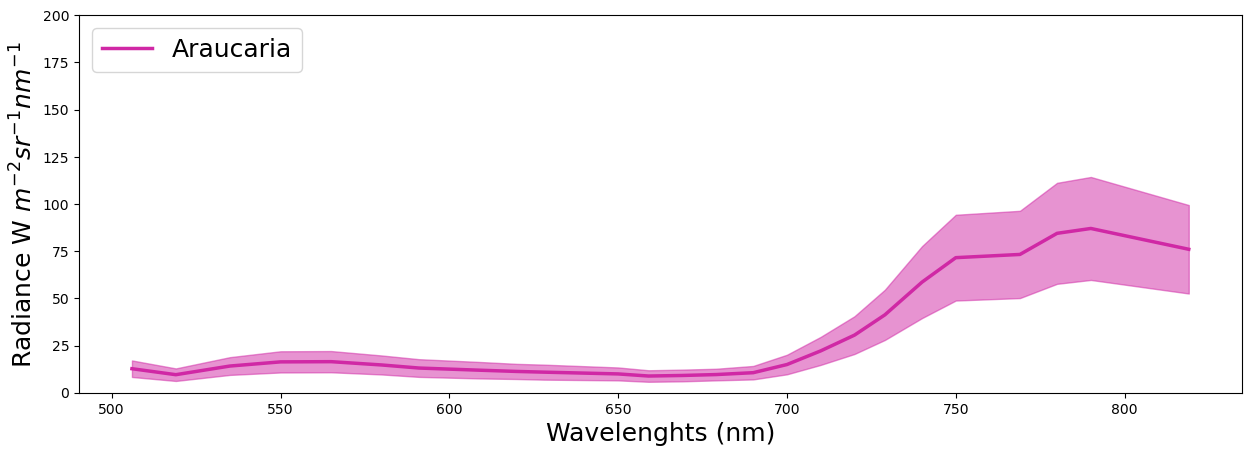}
\end{subfigure}\\
\begin{subfigure}{.5\textwidth}
  \centering
  \includegraphics[width=1\linewidth]{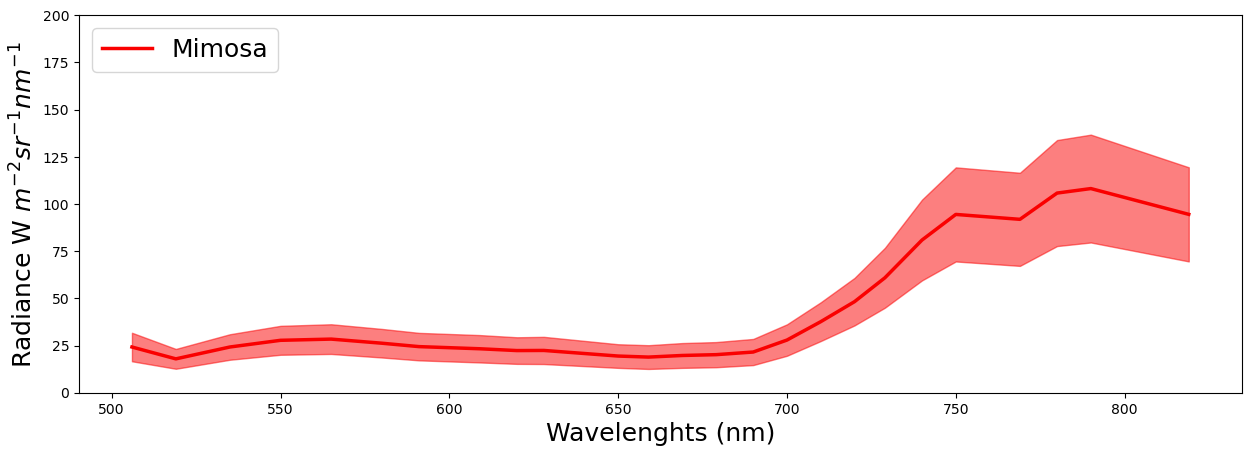}
\end{subfigure}%
\begin{subfigure}{.5\textwidth}
  \centering
  \includegraphics[width=1\linewidth]{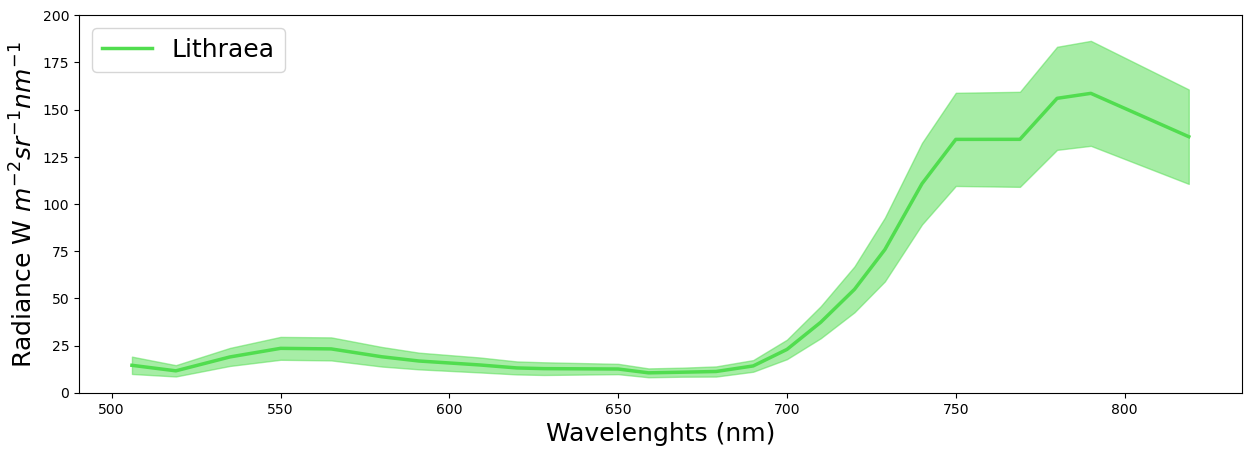}
\end{subfigure}\\
\begin{subfigure}{.5\textwidth}
  \centering
  \includegraphics[width=1\linewidth]{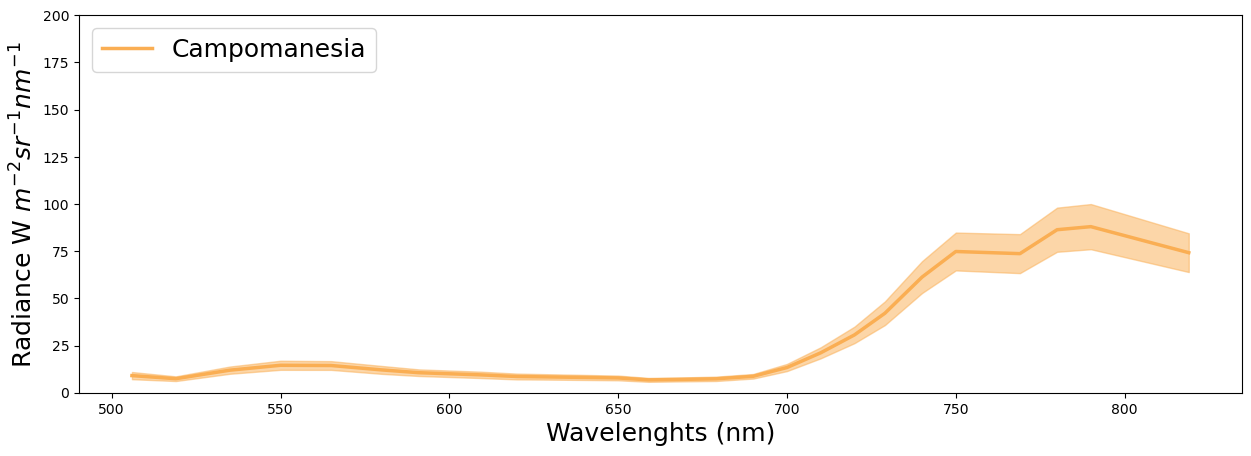}
\end{subfigure}%
\begin{subfigure}{.5\textwidth}
  \centering
  \includegraphics[width=1\linewidth]{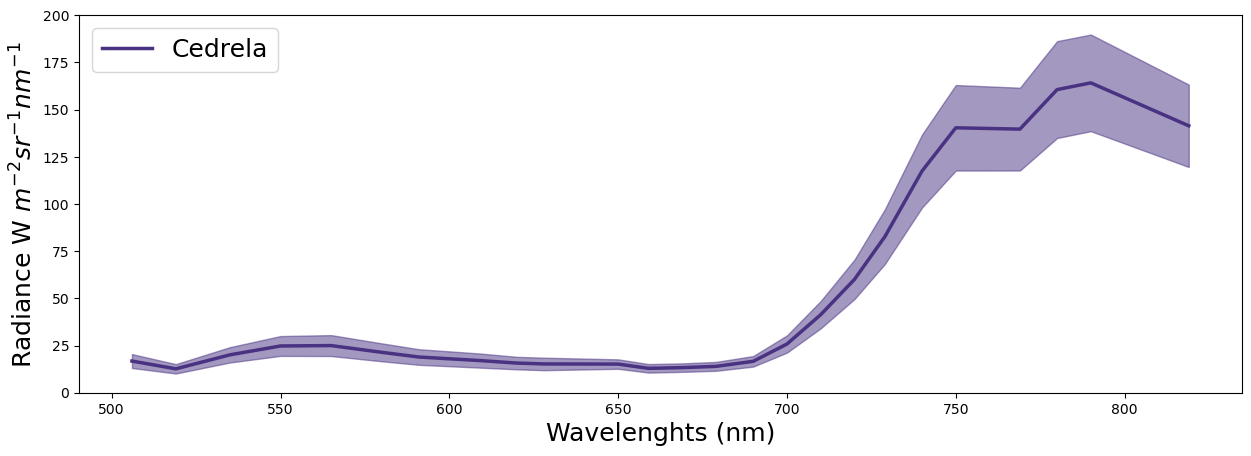}
\end{subfigure}\\
\begin{subfigure}{.5\textwidth}
  \centering
  \includegraphics[width=1\linewidth]{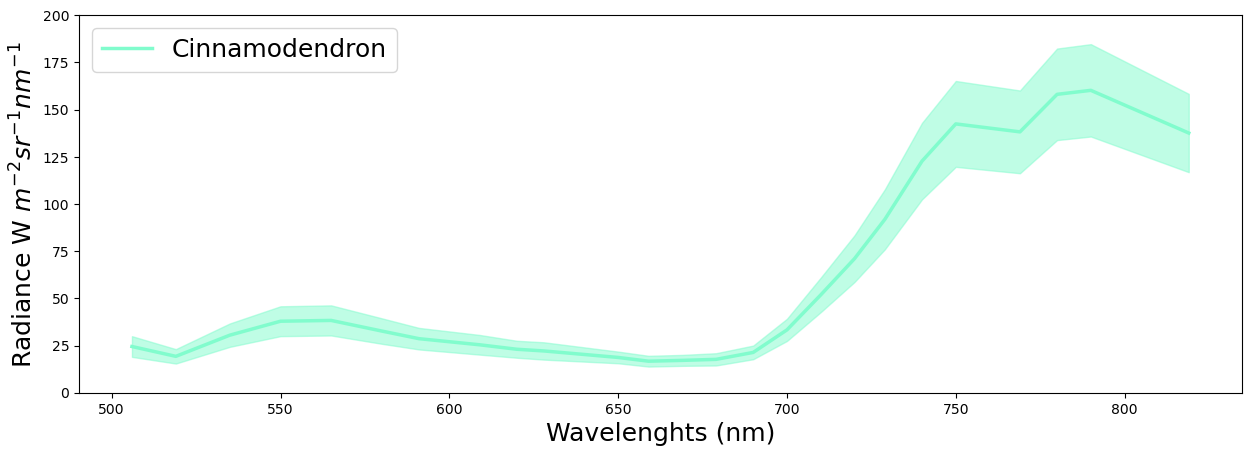}
\end{subfigure}%
\begin{subfigure}{.5\textwidth}
  \centering
  \includegraphics[width=1\linewidth]{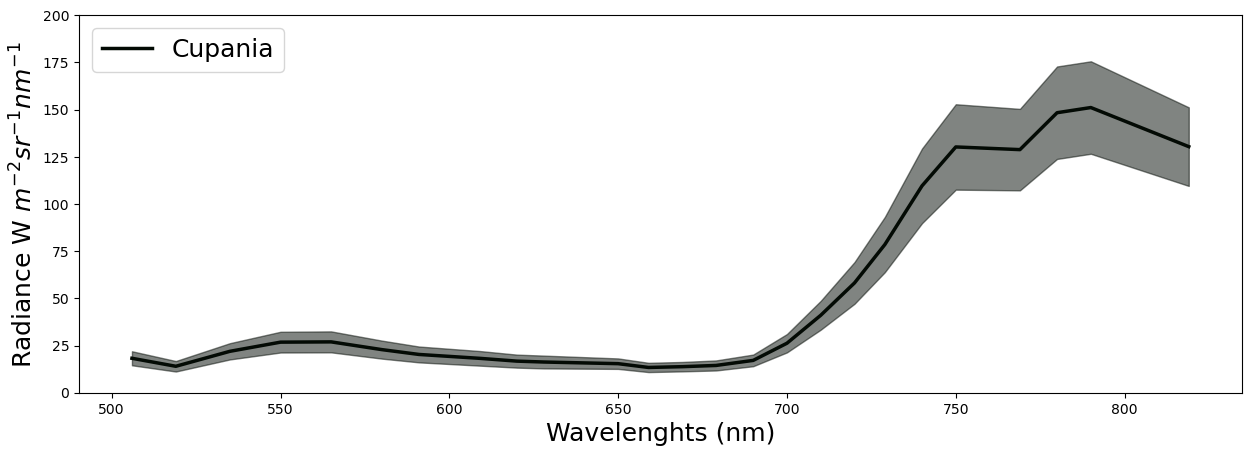}
\end{subfigure}\\
\begin{subfigure}{.5\textwidth}
  \centering
  \includegraphics[width=1\linewidth]{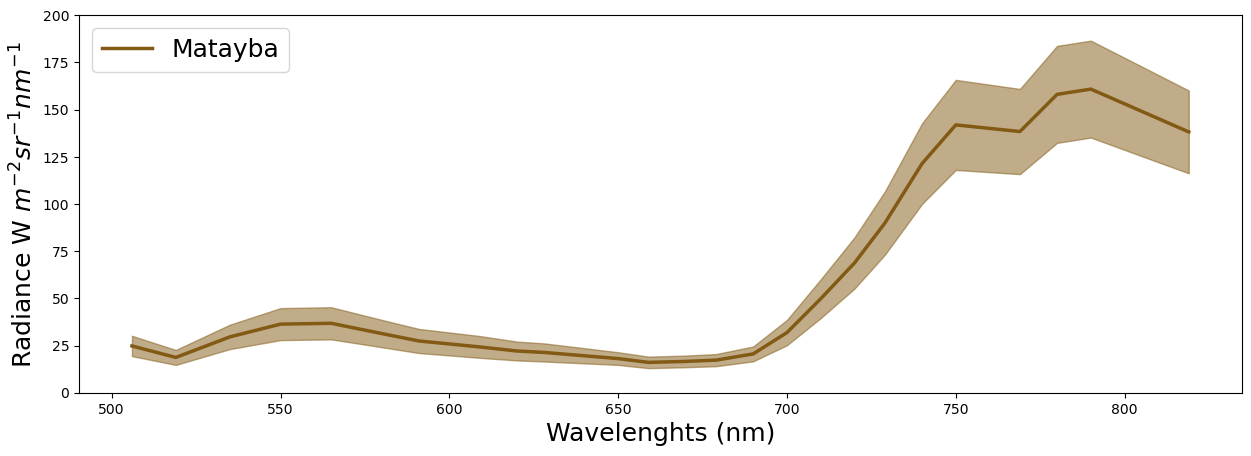}
\end{subfigure}%
\begin{subfigure}{.5\textwidth}
  \centering
  \includegraphics[width=1\linewidth]{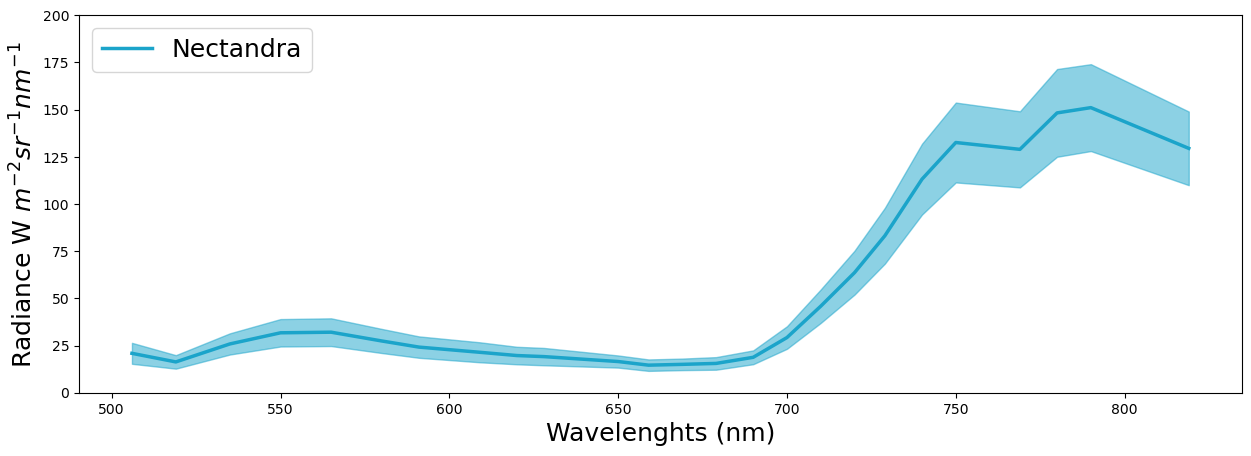}
\end{subfigure}\\
\begin{subfigure}{.5\textwidth}
  \centering
  \includegraphics[width=1\linewidth]{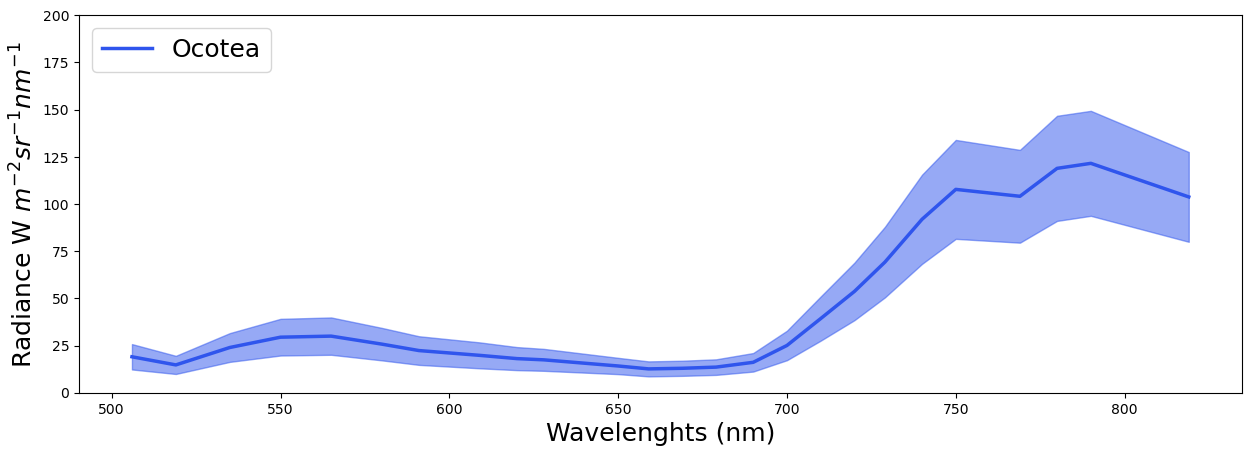}
\end{subfigure}%
\begin{subfigure}{.5\textwidth}
  \centering
  \includegraphics[width=1\linewidth]{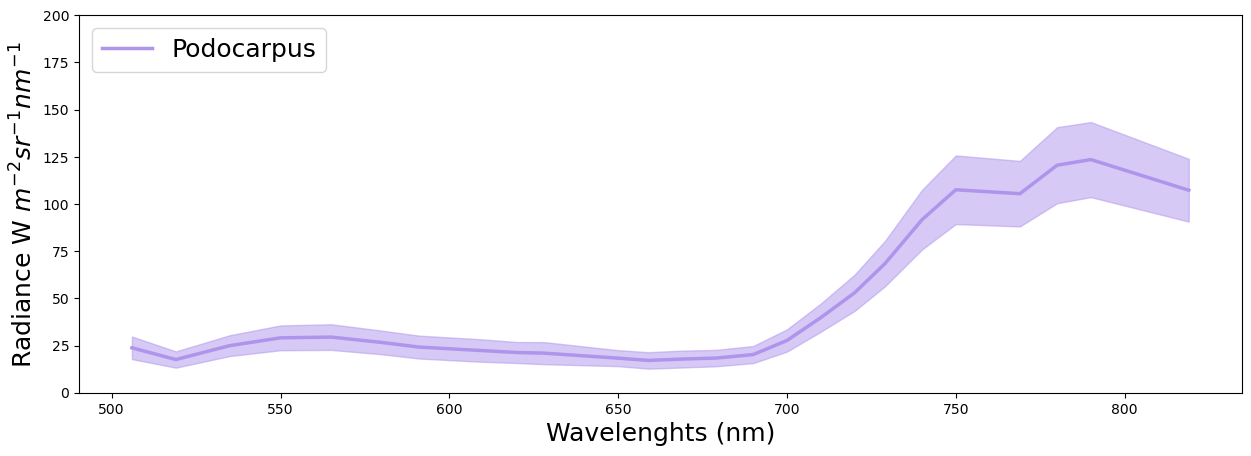}
\end{subfigure}\\
\begin{subfigure}{.5\textwidth}
  \centering
  \includegraphics[width=1\linewidth]{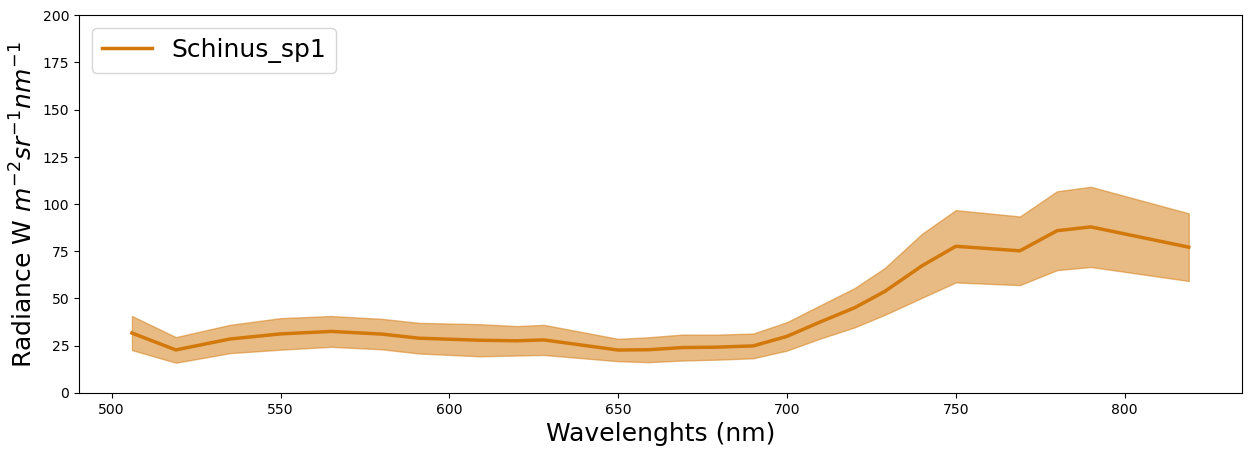}
\end{subfigure}%
\begin{subfigure}{.5\textwidth}
  \centering
  \includegraphics[width=1\linewidth]{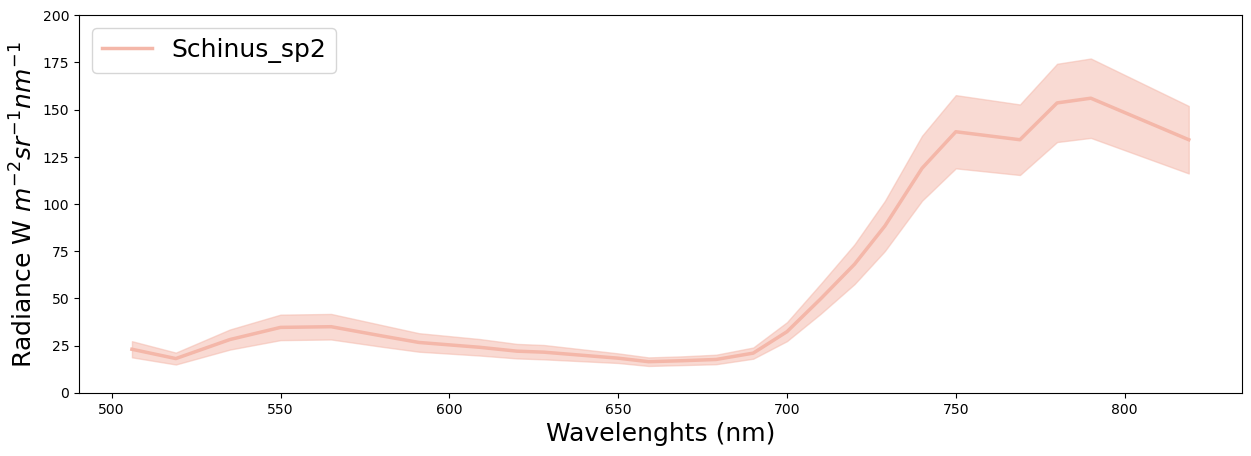}
\end{subfigure}
\caption{Spectral radiance curves for the 14 tree species under investigation.}
\label{fig:spectra}
\end{figure}

The average spectral radiance (with standard deviation) for the 14 tree species under investigation is presented in Figure \ref{fig:spectra}. No relevant difference among the species were observed in the the visible range (506–700 nm). In the green peak region (535 to 580 nm), \textit{Matayba} and \textit{Cinnamodendron} present the greatest mean spectral radiance values. The lowest mean radiance values were observed for \textit{Araucaria} and \textit{Campomanesia}, the later one having also the lowest standard deviation. The difference in spectral radiance values among groups of species becomes more evident in the NIR range (700–819 nm). However, even in this region, the intra-group discrimination remains challenging, for example: \textit{Araucaria}, \textit{Campomanesia} and \textit{Schinus sp1}; and \textit{Podocarpus}, \textit{Mimosa}, and \textit{Ocotea}.

\section{Method}\label{method}

\begin{figure}
	\centering
		\includegraphics[width=0.99\linewidth]{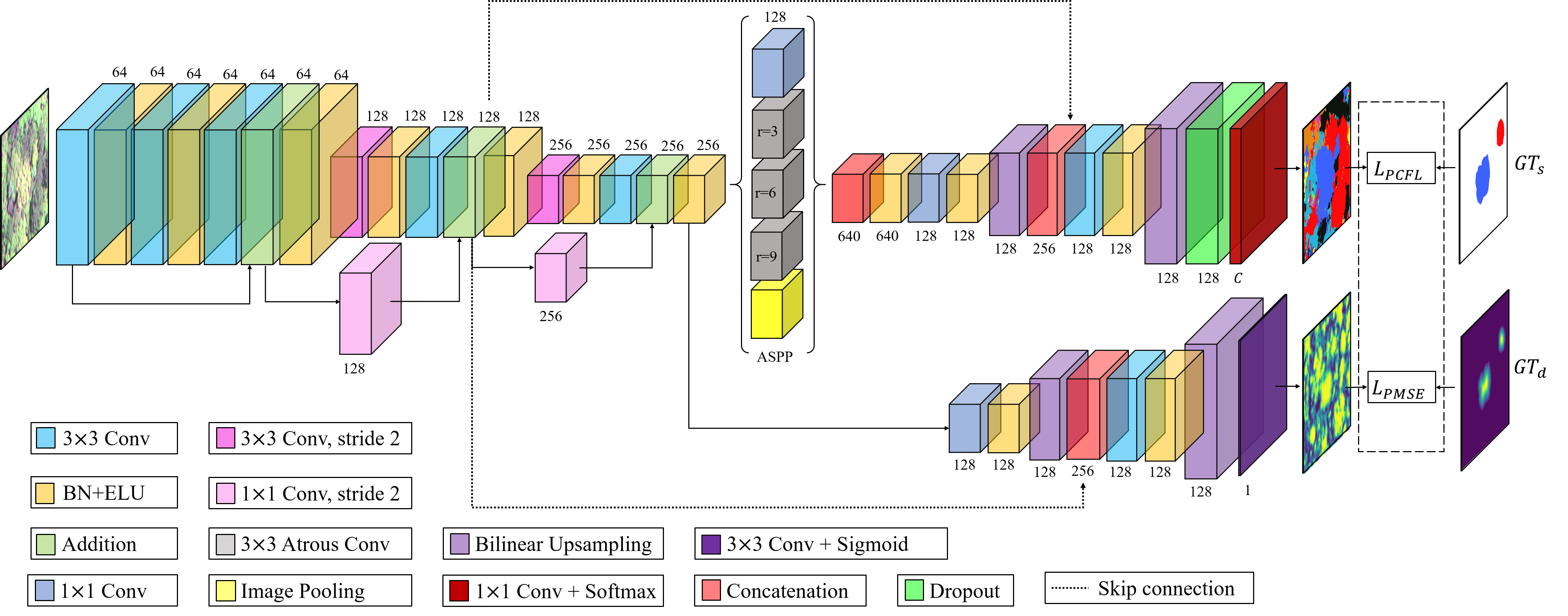}
	\caption{ResNet and DeepLabv3+ based network architecture. $C$ stands for the number of classes, $r$ stands for the atrous rate, and $GT_s$ and $GT_d$ stands for the ground truth and reference distance map, respectively.}
	\label{fig:model}
\end{figure}

Our approach implements a novel multi-task FCN trained with sparse polygon-level annotations (MTFCsp) for dense semantic segmentation of tree species. Our architecture (see Figure \ref{fig:model}) can be viewed as an encoder-focused architecture \citep{crawshaw2020multi}, a type of multi-task learning architecture that learns a generalizable representation in the encoding phase (i.e., backbone network) and task-specific representations in the decoder phase. It takes hyperspectral images and produces two outputs, a class probability map and a distance map, with the input images' resolution. The class probability map holds semantic label posterior probabilities for each pixel in the image, whereas the distance map gives the distance of each pixel within the tree crowns to the closest boundary. 

Since our training set comprises a small number of ITC samples, training solely for the semantic segmentation task would most likely cause over-fitting on the training set and deliver poor performance in the test set. We hypothesize that introducing a distance map regression as a secondary task will act as a regularizer for the semantic segmentation task and ultimately improve its generalization. The use of a distance map estimation as a supplementary task has recently emerged as an effective tool for improving the semantic segmentation in urban regions from RS data \citep{bischke2019multi,diakogiannis2020resunet}.

For our multi-task learning approach, the loss function is defined as the linear combination of the two task-specific losses $L_{total} = L_1 + \lambda L_2$, where $\lambda$ represents a weight parameter. To train the network with sparse annotations, we use the definition of partial loss function for both tasks. Sections \ref{lossfun}, \ref{reggen} and \ref{arquitecture} present a detailed description of the proposed loss functions and the network architecture. 
 
\subsection{Semantic segmentation with sparse polygon-level annotation} \label{lossfun}
Semantic segmentation networks are commonly trained using the cross-entropy loss function between the predicted label $\hat{y}$ and the ground truth label $y$. Given an input image  $I\in {R}^{W \times H}$, the categorical cross-entropy function for a multi-class semantic segmentation problem can be written as follows:
\begin{equation}\label{equ:cross}
	L_{C} = -\frac{1}{|\Omega_L|}\sum_{i\in\Omega_L}^{}\sum_{j\in C}^{} y_{i,j}\log(\hat{y}_{i,j})
\end{equation}
\noindent
where $\Omega_L$, with $|\Omega_L|$ = WH, is the set of labeled pixels, $C$ is the number of classes, $y_{i,j}$ is a binary indicator of pixel $i$ belonging to class $j$, and $\hat{y}_{i,j}$ corresponds to the model’s estimated class probability of pixel $i$ belonging to class $j$. The class probability is commonly calculated by applying the softmax activation function to the network output. 

To effectively train a dense semantic segmentation network with partial annotations, i.e., $|\Omega_L| \neq$ WH, we used the definition of partial loss \citep{wu2018scribble,tang2018normalized}. This type of loss only back-propagates the losses from pixels $i$ that belong to the annotated set $\Omega_L$ and ignores the loss in other pixels. In \citet{wu2018scribble}, the authors found that adopting such a simple modification to the loss function improves results substantially for scarce labeling sets. 

To tackle class imbalance, we use the categorical focal loss \citep{lin2017focal} instead of the cross-entropy loss. The focal loss addresses the class imbalance by forcing the standard cross-entropy to down-weight the contribution of well-classified examples, focusing on those samples difficult to classify. Combining both concepts, the task-specific $L_1$ loss, called partial categorical focal loss ($L_{PCFL}$), can be written as the following:
\begin{equation}\label{equ:pcfl}
	L_{PCFL} = - \frac{1}{|\Omega_L|} \sum_{i \in \Omega}^{} \omega_{i} \sum_{j \in C}^{} y_{ij}\left(1 - \hat{y}_{ij}\right)^\gamma \log(\hat{y}_{ij})
\end{equation}
\noindent
where $\omega_{i} = 1$ if pixel $i \in \Omega_L$ and 0 otherwise, and $\gamma \geq 0$ is the focusing parameter. The focusing parameter weights the loss depending on how well the samples are classified to prioritize hard examples learning. When $\gamma = 0$, $L_{PCFL}$ is equivalent to the partial cross-entropy loss \citep{wu2018scribble}. Notice that in the above definition, $\Omega_L \in \Omega$, with $|\Omega|$ = WH. Figure \ref{fig:model} illustrates the ground truth labeled image ($GT_s$) for an input image with two annotated polygons (red, blue) where the white pixels correspond to the unknown region.

\subsection{Distance map as auxiliary task} \label{reggen}
\begin{figure}
\centering
\begin{subfigure}{.3\textwidth}
  \centering
  \includegraphics[width=.8\linewidth]{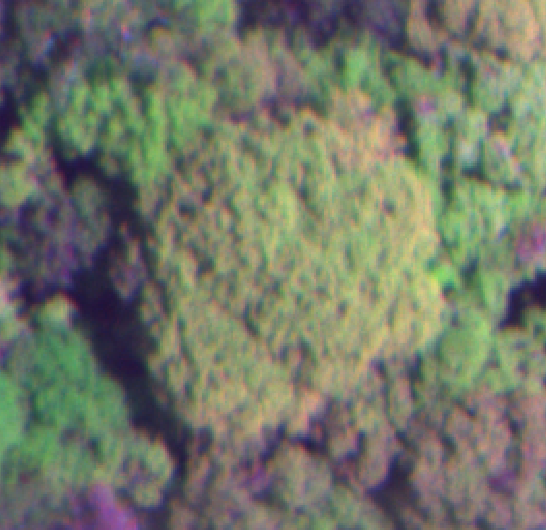}
  \caption{}
  \label{fig:clip}
\end{subfigure}%
\begin{subfigure}{.3\textwidth}
  \centering
  \includegraphics[width=.8\linewidth]{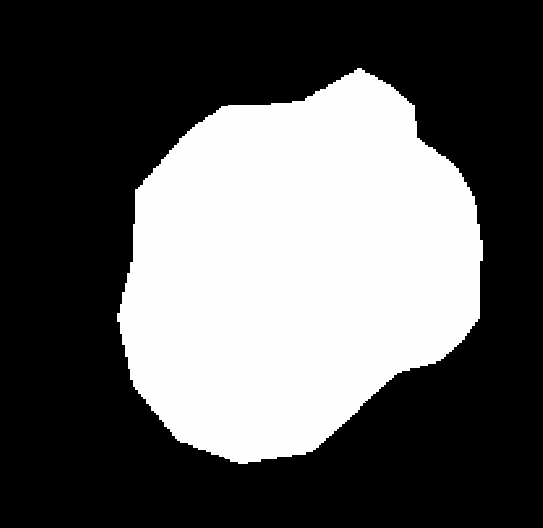}
  \caption{}
  \label{fig:clipb}
\end{subfigure}%
\begin{subfigure}{.3\textwidth}
  \centering
  \includegraphics[width=.8\linewidth]{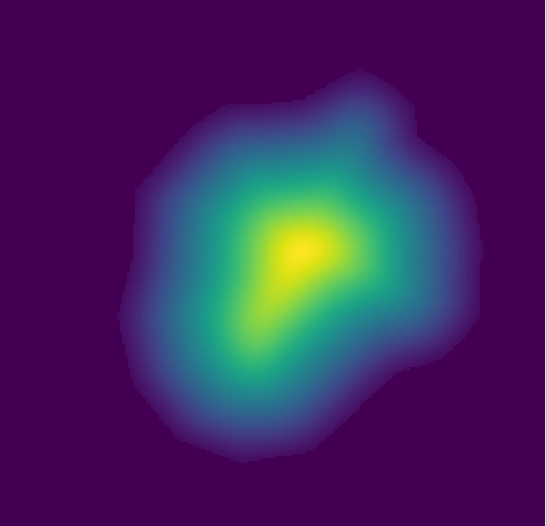}
  \caption{}
  \label{fig:clipd}
\end{subfigure}
\caption{Extracting the distance transform of a single tree. (a) hyperespetral image, (b) binary mask of the ITC sample, (c) distance map.}
\label{fig:edt}
\end{figure}

The distance map estimation auxiliary task aims to improve the segmentation network's generalization. The network will learn a regression function that maps the input image to a distance map where each pixel within the tree crowns holds the distance to the closest crown boundary. We compute such distances from the ITCs reference and train the network using a standard loss for regression. We implement a simple pipeline for creating the training set for distance maps. First, we generate the corresponding binary mask for each annotated polygon. Second, we compute the distance map $d$ for the resulting binary image using the Euclidean Distance Transform (EDT). This operation results in a distance map when each pixel value within the polygons is the Euclidean distance to the closest boundary. Third, we smooth the distance map by applying a 2D Gaussian kernel. Finally, we normalize all distances for each polygon between 0 and 1. Since the target region is highly heterogeneous, with different crown sizes even for the same tree species, we deliberately opted for an ITC-level normalization to ensure that the maximum values for each ITC are 1. Figure \ref{fig:edt} illustrates the distance map result for a single-tree. Note that the peak of the distance map is at the center of the tree crowns. 

Our intuition is that forcing the network to assign low values to the tree crown edges could help identify individual trees, even those of the same semantic class. It is worth noticing that the segmentation branch does not include any restriction about the polygon's edges, and therefore this extra training signal can potentially foster more accurate results.

As in the semantic segmentation branch, we employed a partial loss function to train the distance map estimation branch. We modeled the problem as a regression problem employing the standard Mean Square Error (MSE) as the $L_2$ loss function in our multi-task problem. MSE is the sum of squared distances between the reference map and the estimated map, and our modified version takes the following form:
\begin{equation}\label{equ:pmse}
	L_{PMSE} = - \frac{1}{|\Omega_L|}\sum_{i \in \Omega}^{} \omega_{i} (d_i - \hat{d}_i)^2
\end{equation}
\noindent
where $\omega_{i} = 1$ if pixel $i \in \Omega_L$ and 0 otherwise, $d$ is the reference map and $\hat{d}$ is the network’s estimated distance map.

\subsection{Architecture Design} \label{arquitecture}
Figure \ref{fig:model} depicts the architecture design of our approach. MTFCsp consists of a single shared encoder that learns general low-level features and two task-specific decoders that learn spectral and textural features concerning each specific task. The network is trained in a weakly-supervised end-to-end fashion using the total loss function $L= L_{PCFL} + \lambda L_{PMSE}$, based on the previously described partial loss functions. In our experiments we set $\lambda =1$.

\subsubsection{Proposed shared encoder}
\begin{figure}
	\centering
		\includegraphics[width=0.4\linewidth]{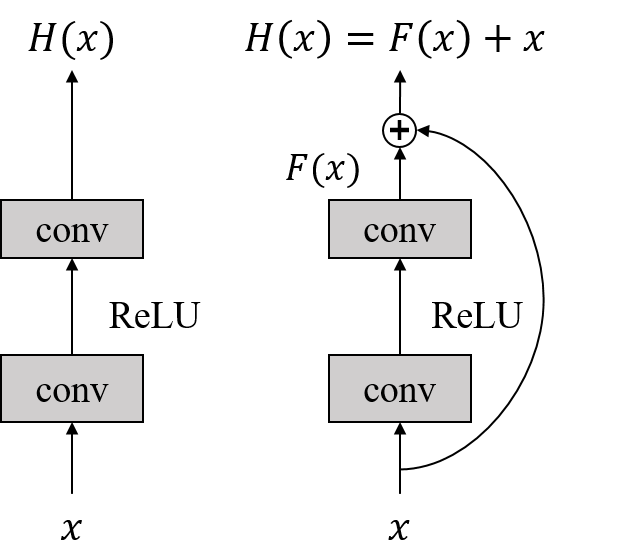}
	\caption{A regular block (left) and a residual block (right).}
	\label{fig:resblock}
\end{figure}

The shared encoder was designed based on the ResNet architecture \citep{he2016deep}. These networks learn not an underlying mapping function $H(x)$ (Figure \ref{fig:resblock} (left)) but a residual function $H(x)-x$ that is expected to be more discriminant. In its final form, the residual block learns a function $H(x)=F(x)+x$ (Figure \ref{fig:resblock} (right)), and uses a shortcut connection that handles the gradient vanishing problem without adding any extra parameter to the network.

Even though residual blocks enable deeper architectures, we experimentally setup a network, called henceforth ResNet-9 (Figure \ref{fig:model} encoder phase), composed of only nine convolutional operations in 3 residual blocks. In an exploratory test, we found that this shallow configuration provided a better trade-off between computational cost and accuracy than standard versions of ResNet, including ResNet-18 and ResNet-50. Each residual block has two 3$\times$3 convolutional layers, and before each convolution within the residual blocks, we apply Batch Normalization (BN) and ELU activation function. The spatial dimension was reduced by using convolution operations with stride 2. As in ResNet architecture, we used a first convolutional layer that did not reduce the spatial dimension. The number of filters doubles periodically at each residual block, and the spatial resolution of the output feature maps is four times smaller than the input resolution.

\subsubsection{Proposed decoder for semantic segmentation}
\begin{figure}
	\centering
		\includegraphics[width=0.7\linewidth]{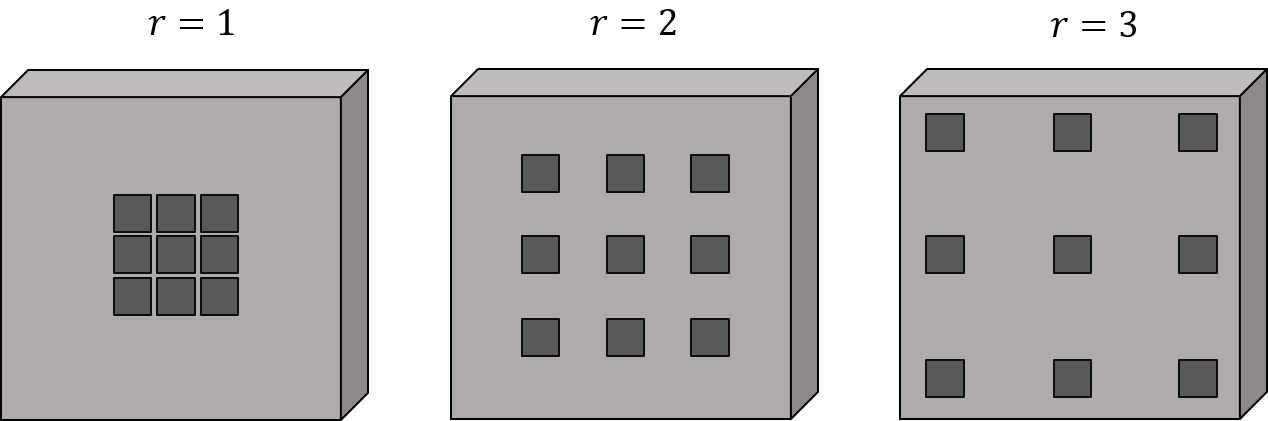}
	\caption{Example of atrous convolutions with different atrous rates.}
	\label{fig:atrous}
\end{figure}

For the semantic segmentation branch, we feed the ResNet-9 output feature map to a decoder based on the DeepLabv3+ architecture \citep{chen2018encoder}, which is considered the state-of-the-art for this task. We first applied the Atrous Spatial Pyramid Pooling (ASPP) module \citep{chen2017deeplab}, which consists of parallel atrous convolutions operations. The atrous convolution's fundamental characteristic is the filters that have $r-1$ rows/columns of zeros separating neighbouring learnable weights, as shown in Figure \ref{fig:atrous}, when $r$ is the atrous rate that determines the minimum distance between two learnable filter weights. With $x$ as the input, the atrous convolution is defined as:
\begin{equation}\label{equ:atrous}
	y[i] = \sum_{k=1}^{K} x[i + r*k]w[x]
\end{equation}
\noindent
where $y[i]$ is the output feature map at pixel $i$, $w$ is the convolutional filter and $r$ is the atrous rate. Notice that when $r=1$, atrous convolution is equivalent to the standard convolution. This type of convolution allows a larger receptive field without increasing the number of parameters or loss in spatial resolution. Performing these operations in parallel permits to capture context at multiple scales. 

Similar to \cite{chen2018encoder}, our ASPP module (see Figure \ref{fig:model}) consists of 5 parallel operations: an image pooling, a 1$\times$1 convolution, and three 3$\times$3 atrous convolution with $r$ equal 3, 6 and 9, respectively. We concatenate the five outputs and apply a BN and an ELU activation function. Then, we used two convolutional blocks (CB) consisting of a 3$\times$3 convolution followed by a BN, an ELU activation function, and a bilinear upsampling to recover the input spatial resolution. We also use skip-connections by concatenating the first CB's output with the corresponding encoder low-level features (depicted in  Figure \ref{fig:model} with the dotted black lines). The last two layers of our decoder network consist of a Dropout layer and a classification layer implemented with a 1$\times$1 convolution with a softmax activation function that delivers the class membership probabilities. We use this probability map to calculate the $L_{PCFL}$ as described in subsection \ref{lossfun}. It is worth pointing out that the errors obtained in the loss function are propagated only through the semantic segmentation branch and the shared encoder. 

\subsubsection{Proposed decoder for distance map estimation}For the distance map branch, we feed the ResNet-9 output feature map to a decoder composed of two CB, which also consist of a 3$\times$3 convolution followed by a BN, an ELU activation function, and a bilinear upsampling. Again, we used skip connections by concatenating the first CB's output with the corresponding low-level features (Figure \ref{fig:model} dotted black line). Finally, we employed a last 3$\times$3 convolutional layer with a sigmoid activation function to obtain a regression map (i.e., the distance map). The output is then used to compute the $L_{PMSE}$ as described in subsection \ref{reggen}, which is propagated only through the regression branch and the shared encoder. In preliminary experiments we found that this configuration provided a good response for the distance map estimation.

\section{Experiment design}\label{exp}
We proposed two experiments in this study. First, we run an experiment to evaluate the partial loss using the segmentation branch uniquely deriving the single-task fully convolutional architecture for sparse annotation (STFCsp). Second, we run the experiment using both branches, i.e., the proposed MTFCsp architecture, and compared their results. This paired experiment's objective is to evaluate the benefits of including the second task as a regularizer.

\begin{figure}
    \centering
    \includegraphics[width=0.9\linewidth]{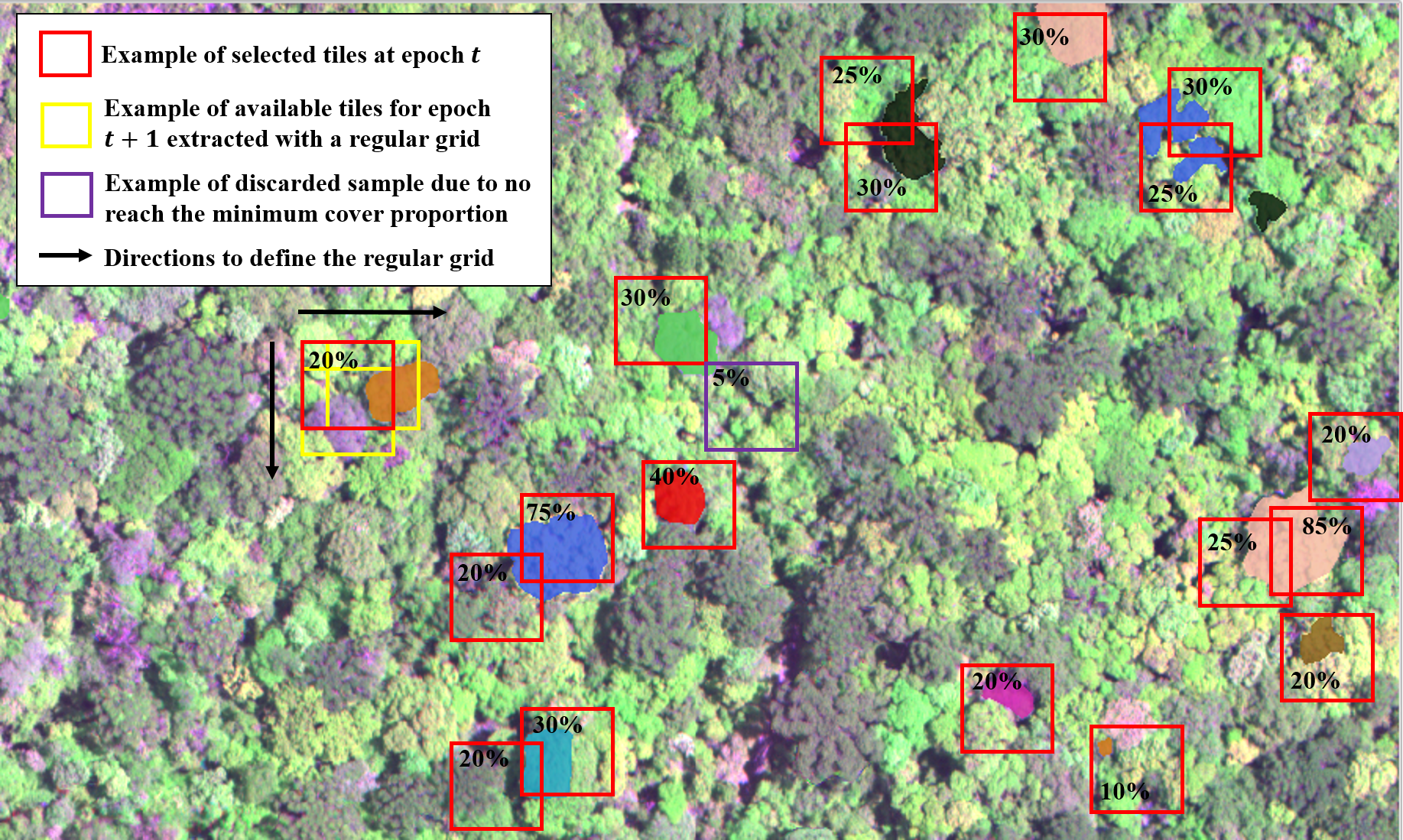}
    \caption{Scheme of the random-crop strategy based on cropping random image tiles guaranteeing at least 10\% of cover proportion of one of the target classes.}
    \label{fig:crop}
\end{figure}

To run our experiments, we first selected 38 of the ITC samples to compose the training set. We trained and validated STFCsp and MTFCsp models using image tiles of size 128$\times$128. To extract the tiles, we used a random-crop strategy (see Figure \ref{fig:crop}) with the following pipeline. Firstly, we used a regular grid to sample the tiles' central coordinates. Since our training set consists of a small number of annotated ITCs, we set the grid spacing to derive 98\% of overlap between neighboring tiles and create sufficient training samples. Secondly, we cropped square tiles of 128$\times$128 from the orthoimage, the digitized polygons, and the distance map (only for the MTFCsp model). Using this strategy, we randomly cropped $140,000$ image tiles on the fly at each epoch, guaranteeing that at least 10\% of each tile is annotated with a reference ITC.

As reported in Table \ref{tab:tree}, some species outnumber others by a large margin in the number of annotated pixels. Class imbalanced datasets significantly reduce deep learning models' accuracy, creating a bias to learn those features that best discriminate among the classes with the higher number of samples. Thus, we proposed a strategy that targets ensuring that all classes have similar probabilities of appearing in a cropped tile by oversampling the under-represented classes. We implemented an image tiles selection process that produces multiple views (total or partial) of the same tree, operating a data augmentation process. Besides, we employed random rotations (90$^{\circ}$, 180$^{\circ}$, 270$^{\circ}$) and horizontal and vertical flips during training to improve generalization. It is worth pointing out that this strategy does not ensure the same number of labeled pixels per class, as annotated trees with higher diameters have more annotated pixels within a tile than trees with smaller diameters. 

The network architecture, as detailed in Figure \ref{fig:model} and subsection \ref{arquitecture}, includes a dropout layer before the classification layer. To deal with overfitting, we set a high dropout rate value of $0.65$ that gave the best segmentation performance. For training, the stochastic gradient descent (SGD) optimizer with a momentum of 0.9 was used with an initial learning rate of $0.1$ and an inverse time decay schedule with a decay rate of $0.1$ every five epochs. For the $L_{PCFL}$ function, we selected  $\gamma$ = 2 as in \citep{lin2017focal}. We run each experiment 25 times, using 25 different random seeds to split the available training tiles into training and validation sets. For the validation set, we randomly selected a 1\% hold-off of the training tiles at each run. We trained the models for ten epochs, with eight image tiles per batch. We used ten epochs since both models converged before the 10th epoch. STFCsp usually converged before the 3rd epoch, whereas the MTFCsp model converged at the first five epochs. This behavior was not unexpected since we used heavily overlapped tiles, which implies similar validation and training sets. Even so, we monitored the average F1 score and applied early stop when no improvements higher than 0.9e-4 in the validation set happened in a sequence of five epochs.

At inference time, we applied the trained network to overlapping image tiles using a sliding window strategy and keeping the patch's central region, minimizing border effects. In preliminary experiments, we found that averaging the prediction outcomes considering different overlapping rates improved the result. Hence, we generated classification outputs for 10\%, 30\%, and 50\% overlap and took the average as the final outcome. Finally, we concatenated the outputs to obtain an outcome with the input orthoimage dimensions.

Following our previous work \citep{sothe2020comparative}, we used RF, SVM, and a CNN patch-based (CNN) as the baseline methods. All the three methods received as input the 25 bands and used the same parameters configuration and training procedure reported in \citet{sothe2020comparative} for the VNIR dataset. The size of the input patch/tile for the CNN network was set to 33 $\times$ 33. For more details about the CNN architecture refer to \citep{sothe2020comparative}. We used the same training and test samples used to train and evaluate the STFCsp and MTFCsp methods.

Model performance was evaluated in the annotated test ITCs not used during training. Overall accuracy (OA), Kappa score, user's accuracy (UA), producer's accuracy (PA) and F1 score were computed as performance metrics for the models. The experiments ran on a Linux workstation (Mint 19.2 Cinnamon) with an Intel Core i7-4790, 32Gb RAM, and an NVIDIA GeForce Titan GTX 1080 graphics card (11Gb RAM). The processing chain was implemented on the Python platform using the Tensorflow library.

\section{Results}\label{results}

\subsection{Models performance}
\begin{figure}[t]
\centering
\includegraphics[width=0.8\linewidth]{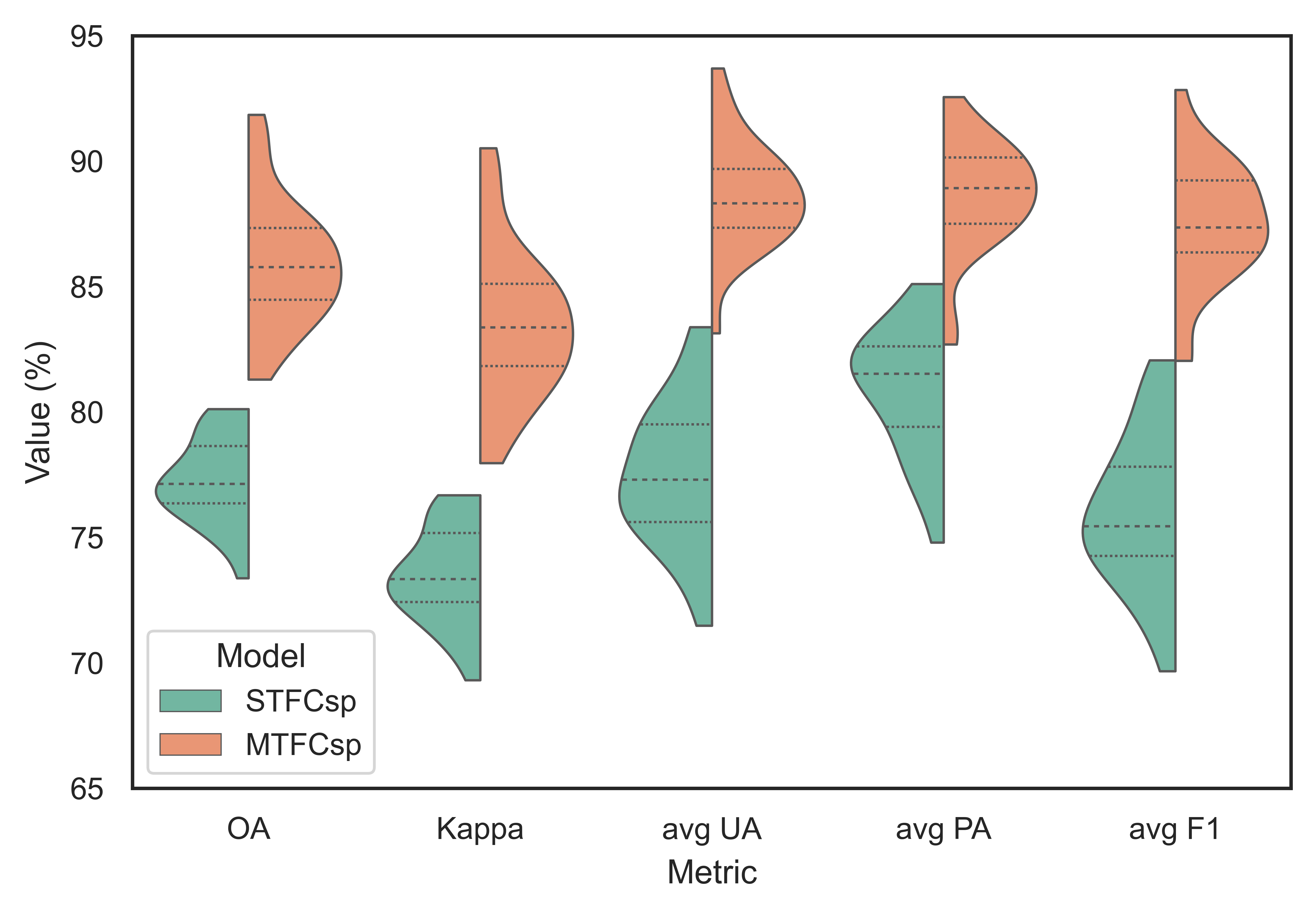}
\caption{Overall accuracy (OA), Kappa, average user's accuracy (avg UA), average producer's accuracy (avg PA) and average F1 score (avg F1) for both models displayed as violin plots. Dotted lines give the upper and lower quartiles and the dashed line gives the median. The stretch of the violins varies according to the variability of the metrics.}
\label{fig:overall}
\end{figure}

Results showed that the multi-task approach MTFCsp outperformed the single-task model STFCsp for all analyzed metrics (see Table \ref{tab:metrics} and Figure \ref{fig:overall}). The performance for the MTFCsp model ranged from 83.50\% (mean Kappa score) to 88.63\% (mean UA), which is significantly superior to STFCsp, with values ranging from 73.51\% (mean Kappa score) to 80.95\% (mean PA); among the 25 realizations. This represents an increase of up to 11\% for user's accuracy and F1 score, indicating that adding the auxiliary task was beneficial over the single-task method. However, we found that for both methods, the performance varied when repeating the training and classification procedures 25 times with random initialization for training and validation samples (for all the 25 realizations, the test set remained the same). The OA of MTFCsp varied between 81\% and 92\%, and the OA of STFCsp varied between 73\% and 80\%. We observed similar variations for Kappa index. Conversely, STFCsp method presents higher variations for the other three metrics, reaching up to 12.4\% for the F1 score.

Compared to the baseline CNN approach, which reached 83.43\% and 80.84\% for OA and Kappa score, respectively, we observed that STFCsp presented a performance drop, with 77.29\%  for OA and 73.51\% for Kappa score (Table \ref{tab:metrics}). However, in terms of UA, PA and F1 score, the STFCsp reports values similar to the CNN method, ranging from 76\% to 80\%. Nevertheless, STFCsp achieved superior performance compared to the SVM and RF baseline approaches. These results indicates the benefits of using the partial loss function to train a fully convolutional approach with scarce and sparse annotated tiles. In addition, this method is less computationally demanding than the CNN baseline (see Table \ref{tab:time}).

\begin{table}[h]
\caption{Mean values for each analyzed metric for both STFCsp and MTFCsp methods; and the results for the baseline methods.}\label{tab:metrics}
\begin{adjustbox}{width=0.65\textwidth,center}
\begin{tabular}{lccccc}
\toprule
Metric & STFCsp & MTFCsp & CNN & SVM & RF\\
\midrule
PA & 80.95 & \textbf{88.59} & 77.35 & 49.97 & 47.75\\
UA & 77.58 & \textbf{88.63} & 77.14 & 53.90 & 54.05\\
F1 & 76.07 & \textbf{87.51} & 75.59 & 49.08 & 47.29\\
OA & 77.29 & \textbf{85.91} & 83.43 & 63.44 & 63.02\\
Kappa & 73.51 & \textbf{83.50} & 80.84 & 56.80 & 56.00\\
\bottomrule
\end{tabular}
\end{adjustbox}
\end{table}

MTFCsp model outperformed by large the STFCsp model for almost all classes according to accuracy metrics (Table \ref{tab:metrics2}). For example, an improvement of more than 10\% in producer's accuracy (PA) was verified for \textit{Matayba} (27.7\%), \textit{Cedrela} (27.6\%), \textit{Schinus sp2} (23.8), and \textit{Podocarpus} (15.5\%). More remarkable improvements were observed in user's accuracy (UA), with an increase of more than 10\% for \textit{Schinus sp2} (42.1\%), \textit{Lithraea} (24.2\%), \textit{Nectandra} (17.7\%), \textit{Cupania} (16.8\%), \textit{Cinnamodendron} (14.5\%) and \textit{Lithraea} (13.54\%). However, we also observed that the STFCsp method outperformed the MTFCsp method for some of the species. \textit{Araucaria}, \textit{Campomanesia}, \textit{Cinnamodendron}, and \textit{Cupania} had accuracy values approximately up to 1.4\% higher, while \textit{Mimosa} achieved 4.3\% of improvement using the STFCsp method. Conversely, in term of UA, STFCsp outperform MTFCsp for only two species, \textit{Araucaria} and \textit{Matayba} up to 2.5\%. In terms of F1 score, the STFCsp method outperformed the multi-task method in up to 1.63\% for \textit{Araucaria} and \textit{Mimosa}.

\begin{table}[!ht]
\caption{Average user's accuracy (UA), producer's accuracy (PA) and F1 score values per class for the 25 realizations.}\label{tab:metrics2}
\begin{adjustbox}{width=0.85\textwidth,center}
\begin{tabular}{lccccccc}
\toprule
\multirow{2}{*}{Species name} & \multicolumn{3}{c}{STFCsp (\%)} & \multicolumn{1}{l}{} & \multicolumn{3}{c}{MTFCsp (\%)} \\ \cline{2-4} \cline{6-8} 
 & \multicolumn{1}{c}{PA} & \multicolumn{1}{c}{UA} & \multicolumn{1}{c}{F1} & \multicolumn{1}{l}{} & \multicolumn{1}{c}{PA} & \multicolumn{1}{c}{UA} & \multicolumn{1}{c}{F1}  \\
\midrule
\textit{Luehea}          & 84.8 & 90.9 & 87.2 && \textbf{88.5} & \textbf{93.8} & \textbf{90.4}\\
\textit{Araucaria}      & \textbf{84.1} & \textbf{90.9} & \textbf{87.1} && 83.3 & 89.4 & 85.5\\
\textit{Mimosa}         & \textbf{94.5} & 96.9 &\textbf{95.4} && 90.2 & \textbf{99.3} & 94.0 \\
\textit{Lithraea}        & 75.3 & 67.5 & 70.6 && \textbf{78.0} & \textbf{91.7} & \textbf{84.1} \\
\textit{Campomanesia}   & \textbf{99.8} & 70.4 & 81.9 && 98.6 & \textbf{76.8} & \textbf{86.0} \\
\textit{Cedrela}        & 69.9 & 91.2 & 78.5 && \textbf{97.5} & \textbf{92.9} & \textbf{95.1} \\
\textit{Cinnamodendron} & \textbf{100} & 77.5 & 85.8 && 99.8 & \textbf{92.0} & \textbf{95.5} \\
\textit{Cupania}        & \textbf{99.9} & 76.1 & 86.0 && 99.7 & \textbf{92.9} & \textbf{96.0} \\
\textit{Matayba}        & 34.0 & \textbf{92.0} & 49.4 && \textbf{61.7} & 89.5 & \textbf{72.2} \\
\textit{Nectandra}      & 71.6 & 73.4 & 71.9 && \textbf{77.4} & \textbf{91.1} & \textbf{83.4} \\
\textit{Ocotea}         & 90.5 & 72.8 & 80.6 && \textbf{94.3} & \textbf{78.7} & \textbf{85.7} \\
\textit{Podocarpus}     & 66.0 & 90.1 & 75.2 && \textbf{81.5} & \textbf{89.5} & \textbf{84.7} \\
\textit{Schinus sp1}    & 89.3 & 70.9 & 78.7 && \textbf{92.6} & \textbf{95.6} & \textbf{94.0} \\
\textit{Schinus sp2}    & 73.6 & 25.6 & 36.7 && \textbf{97.4} & \textbf{67.7} & \textbf{68.6} \\
\bottomrule
\end{tabular}
\end{adjustbox}
\end{table}

\begin{figure}
\centering
\begin{subfigure}{.85\textwidth}
  \includegraphics[width=.99\linewidth]{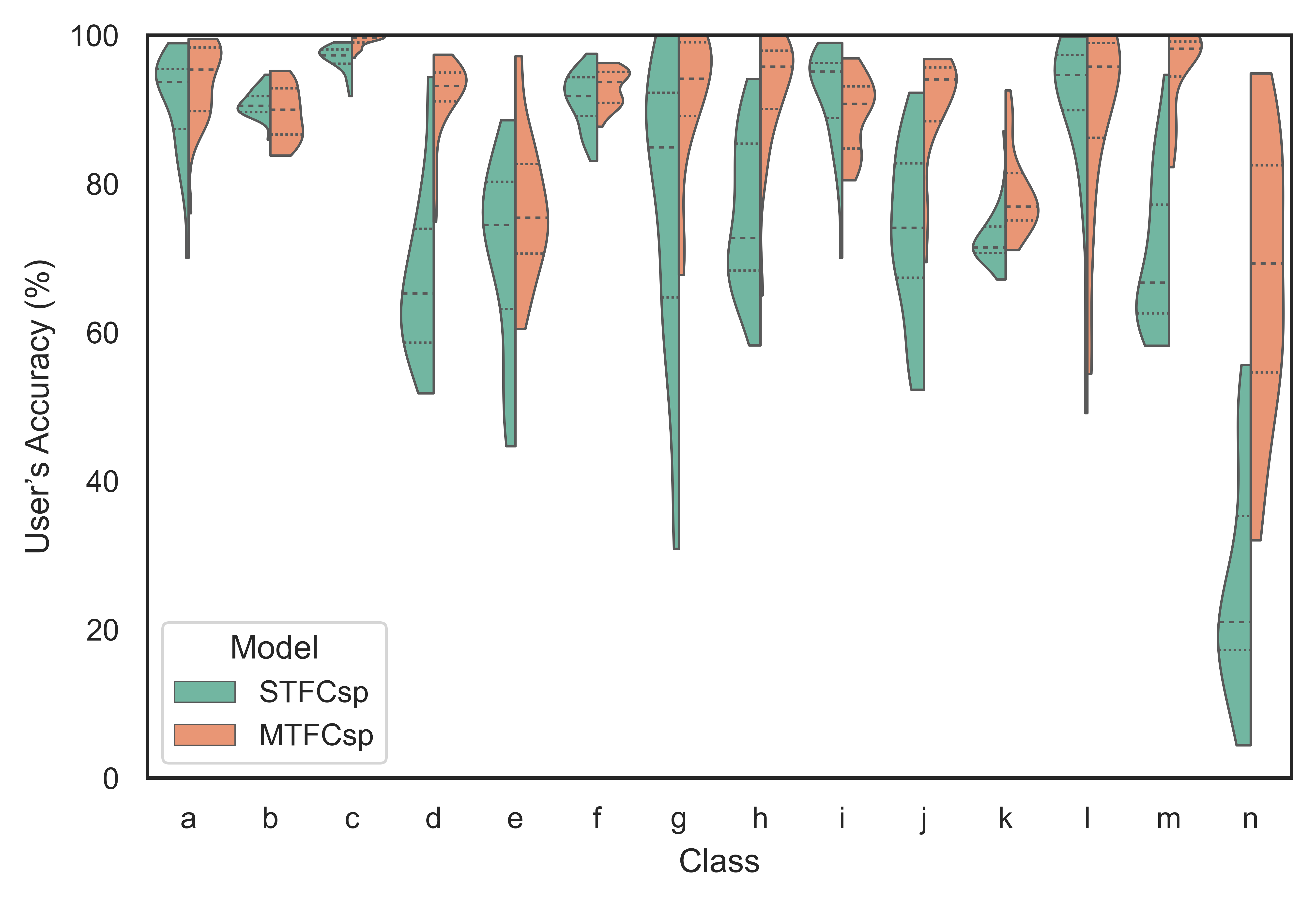}
  \label{fig:sfig3}
\end{subfigure}\\
\begin{subfigure}{.85\textwidth}
  \includegraphics[width=.99\linewidth]{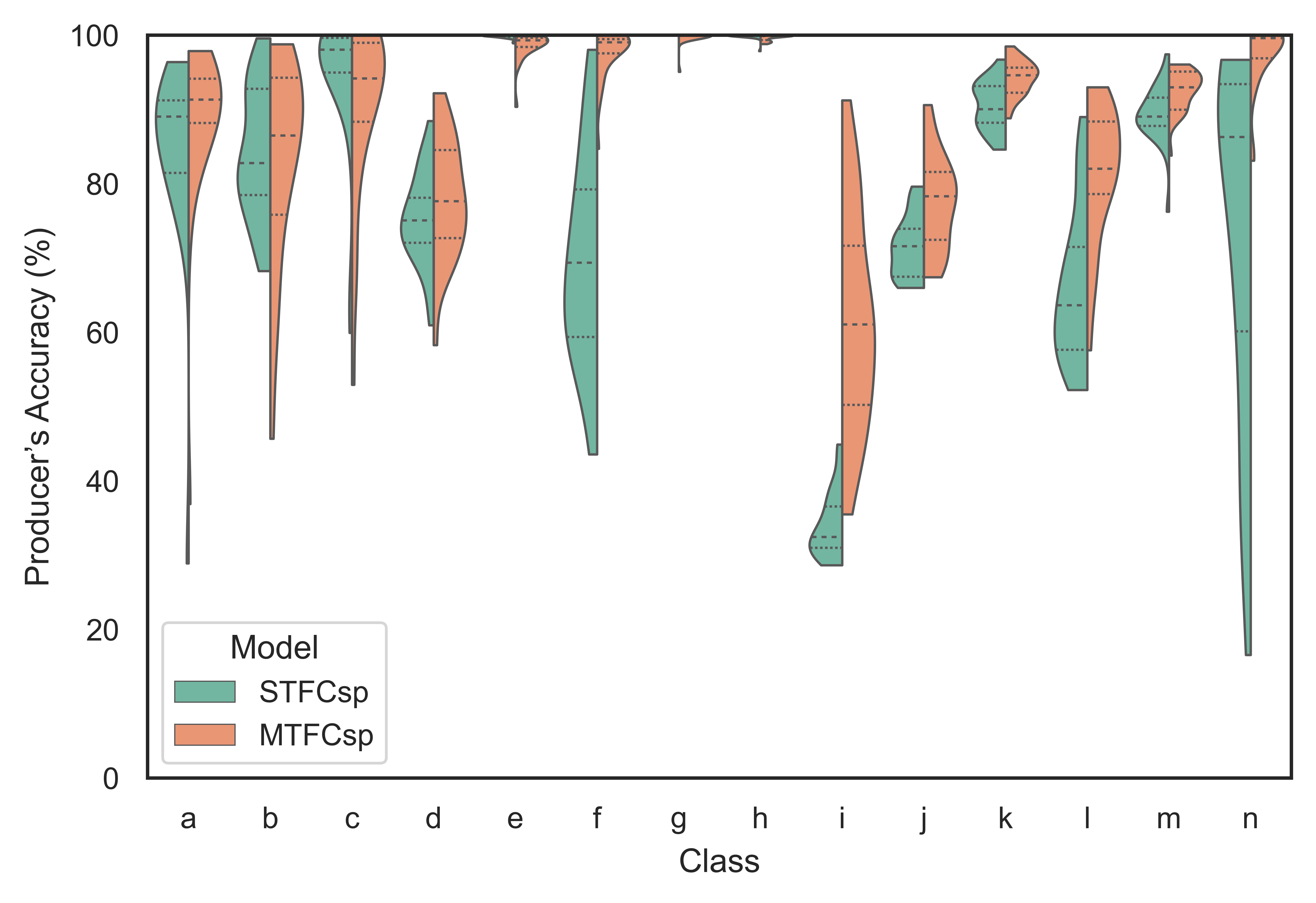}
  \label{fig:sfig4}
\end{subfigure}
\caption{User's accuracy (top) and producer's accuracy (bottom) per class for both models displayed as violin plots. Dotted lines give the upper and lower quartiles, and the dashed line gives the median. The stretch of the violins varies according to the variability of the metrics. Note: Species ID according to Table \ref{tab:tree}.}
\label{fig:perclass}
\end{figure}

In addition, we report that the variability is different among species and methods (Figure \ref{fig:perclass}). For the STFCsp model, we observed a higher variability for \textit{Cedrela} and \textit{Schinus sp2} in terms of PA, and for \textit{Schinus sp2} and \textit{Cinnamodendron} in terms of UA values (Figure \ref{fig:perclass}). In contrast, for MTFCsp model, \textit{Matayba} presents the highest variability in PA, followed by \textit{Schinus sp2} with important variations in UA. These results might be a consequence of the varying number of samples per species and pixels per crown. \textit{Schinus sp2} species has only one ITC with $1,001$ pixels to test the models (see Table \ref{tab:tree}) and, therefore, any miss-classification can cause a significant variation in the performance metrics. Analyzing \textit{Matayba}, one observes that it has the highest number of annotated pixels but also has greater irregularity in their crowns shape and sizes, which can affect the MTFCsp model performance and result in larger variability in the semantic segmentation outcome.

\begin{figure}[!ht]
\begin{subfigure}{.5\textwidth}
  \centering
  \caption{}
  \includegraphics[width=.98\linewidth]{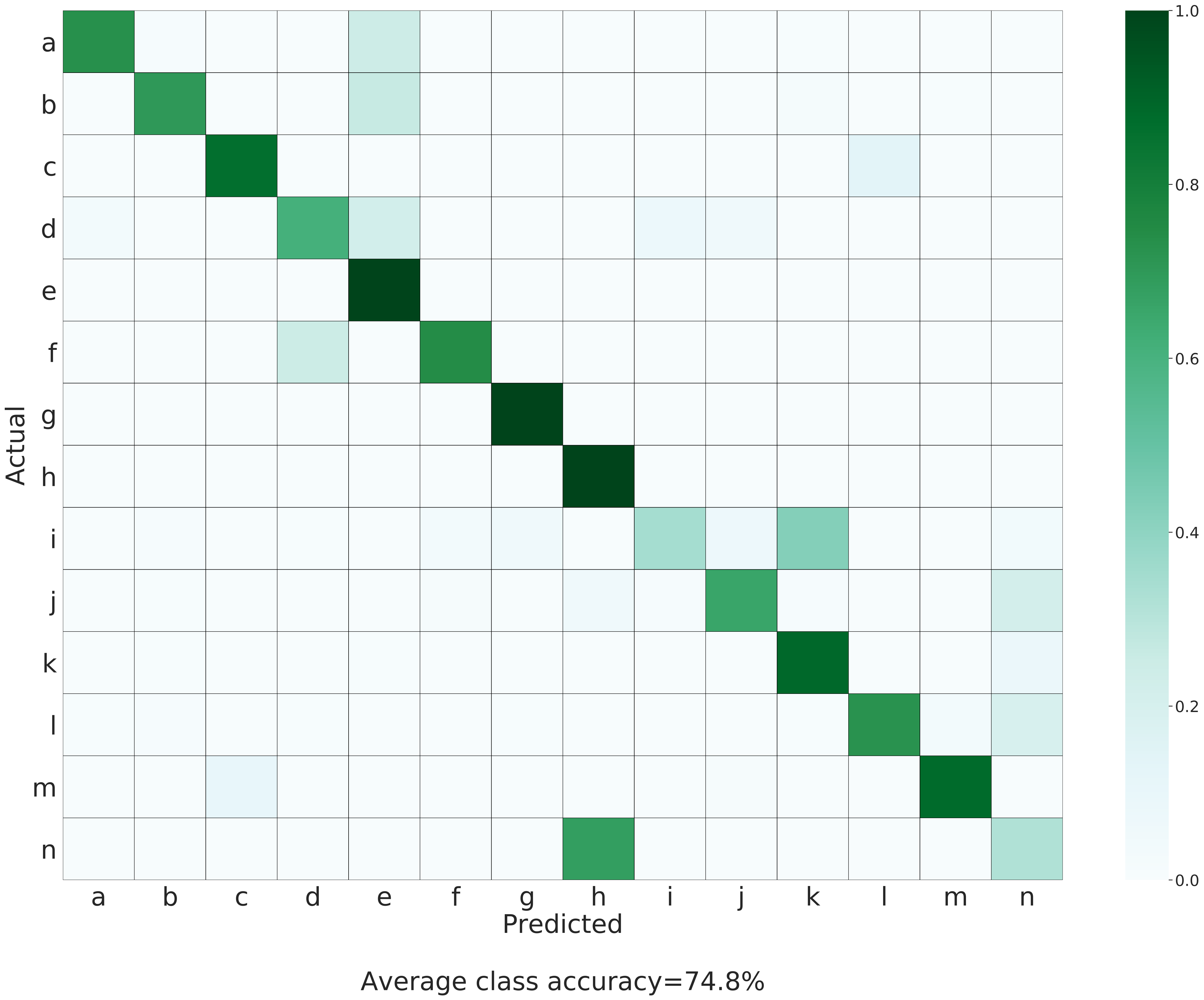}
  \label{fig:lst}
\end{subfigure}%
\begin{subfigure}{.5\textwidth}
  \centering
  \caption{}
  \includegraphics[width=.98\linewidth]{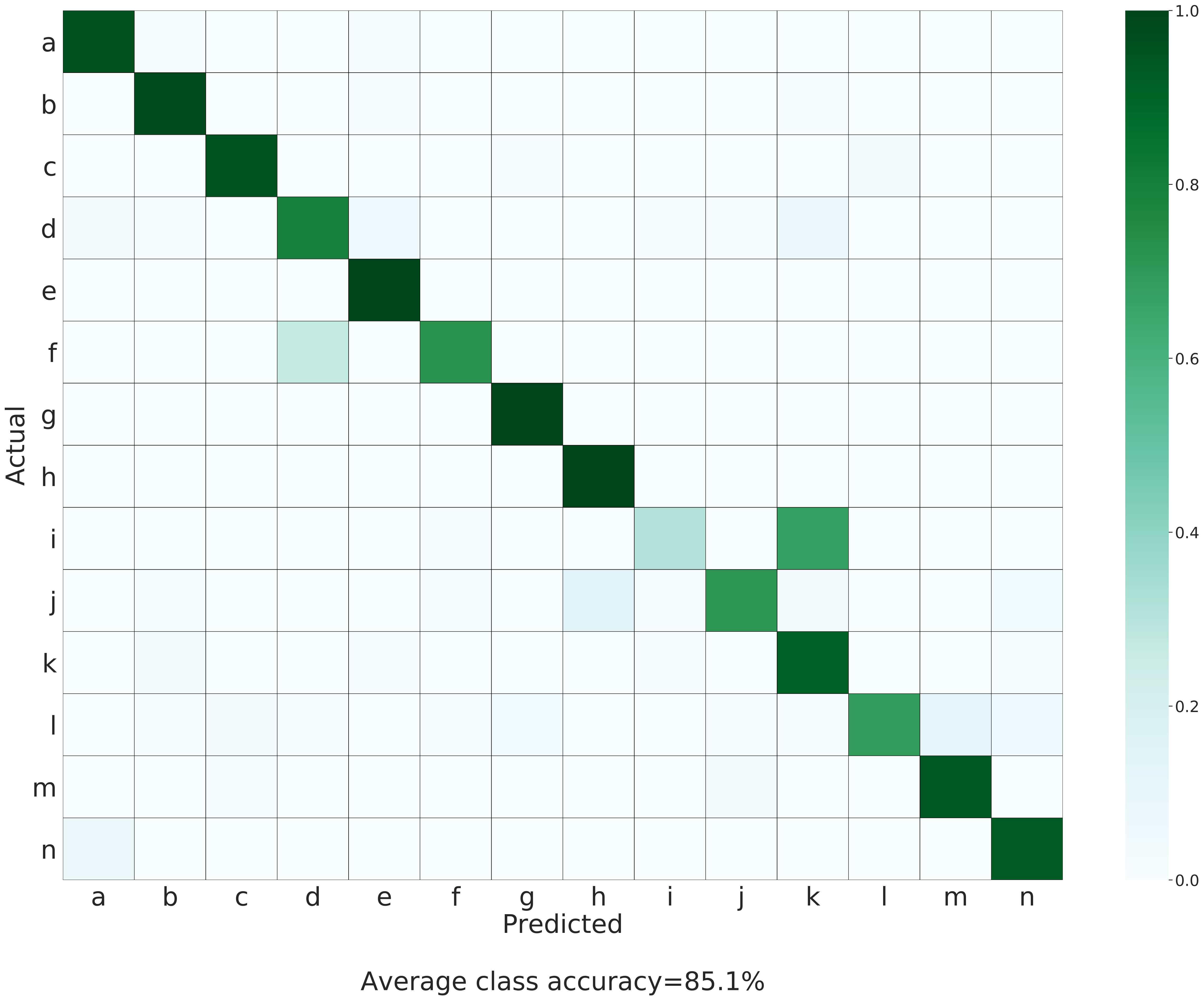}
  \label{fig:hst}
\end{subfigure}\\
\begin{subfigure}{.5\textwidth}
  \centering
  \caption{}
  \includegraphics[width=.98\linewidth]{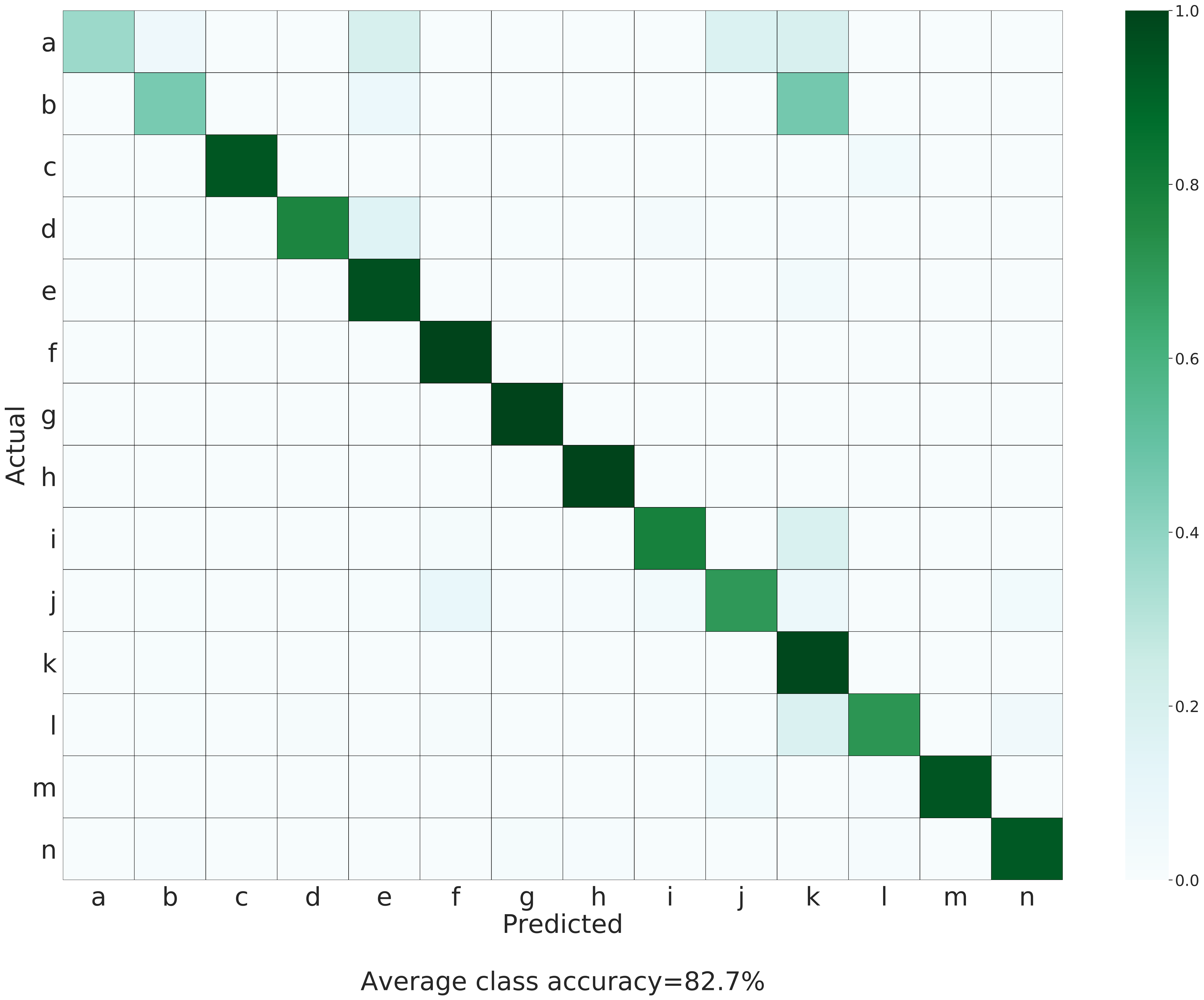}
  \label{fig:lmt}
\end{subfigure}%
\begin{subfigure}{.5\textwidth}
  \centering
  \caption{}
  \includegraphics[width=.98\linewidth]{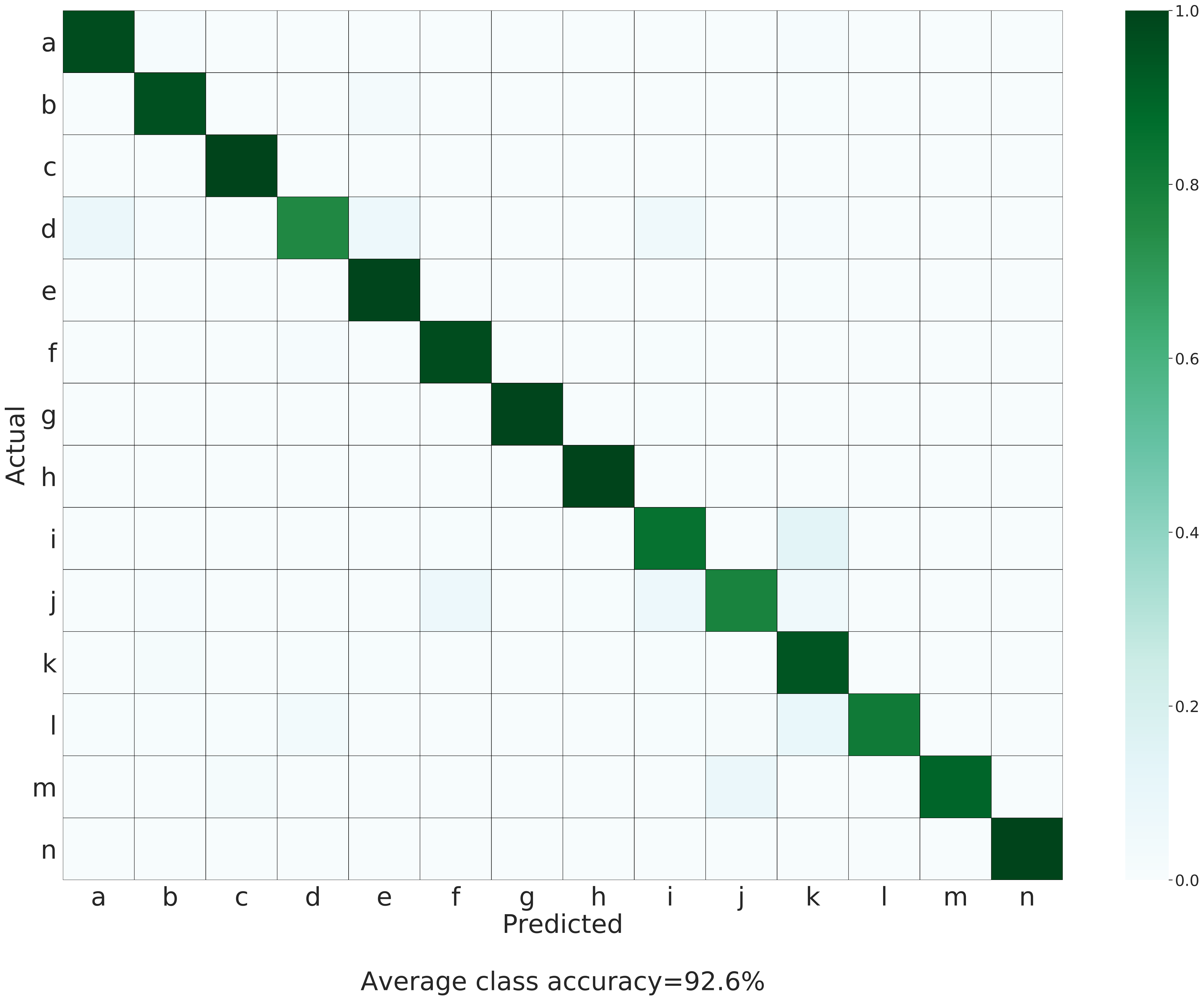}
  \label{fig:hmt}
\end{subfigure}
\caption{Heatmap of the normalized confusion matrices of the models with the lowest and highest performance using the STFCsp ((a) and (b)) and the MTFCsp ((c) and (d)) models evaluated in the test set. Note: Species ID according to Table \ref{tab:tree}.}
\label{fig:conf}
\end{figure}

Figure \ref{fig:conf} shows the heatmap of the normalized confusion matrices yielded by the models with the highest and lowest performance among the 25 realizations. For the STFCsp method, the worst and best realization results reached an average class accuracy (AA) of 74.8\% and 85.1\%, respectively, and for the MTFCsp method, they reached 82.7\% and 92.6\%, respectively.

Considering the STFCsp model, one observes the highest error associated with the \textit{Matayba} class, with accuracy values under 35\% for both the confusion matrices (\ref{fig:lst} and \ref{fig:hst}), and the \textit{Schinus sp2} class, with an accuracy of 32\% for the model with the lowest performance (\ref{fig:lst}). Others species also presented relatively low accuracy values, with values ranging from 60\% to 75\% for \textit{Luehea}, \textit{Araucaria}, \textit{Lithraea}, \textit{Cedrela}, \textit{Nectandra} and \textit{Podocarpus}. We reported the highest misclassification values ($>$ 24\%) for \textit{Schinus sp2} $\times$ \textit{Cupania}, \textit{Matayba} $\times$ \textit{Ocotea}, \textit{Araucaria} $\times$ \textit{Campomanesia}, \textit{Cedrela} $\times$ \textit{Lithraea}, \textit{Luehea} $\times$ \textit{Campomanesia}, \textit{Nectandra} $\times$ \textit{Schinus sp2}, and \textit{Podocarpus} $\times$ \textit{Schinus sp2}. \textit{Campomanesia}, \textit{Cinnamodendron}, \textit{Cupania}, \textit{Ocotea} and \textit{Schinus sp1} were the best classified species, reaching accuracy values above 88\% for both matrices. The high difference in the classification accuracy for the \textit{Schinus sp1} class when comparing both matrices explains the variability observed in the violin-plots.  

Considering the MTFCsp model, one observes the number of errors associated with the \textit{Luehea} and \textit{Araucaria} classes with 36.9\% and 45,7\% of accuracy, respectively, followed by \textit{Nectandra} and \textit{Podocarpus} with values around 70\% for the realization with the lowest average accuracy (Figure \ref{fig:lmt}). The rest of the species presented accuracy values above 77\%. High misclassification rates were observed between \textit{Araucaria} $\times$ \textit{Ocotea} (46.6\%), \textit{Matayba} $\times$ \textit{Ocotea} (18.5\%), \textit{Luehea} $\times$ \textit{Campomanesia} (19.7\%), \textit{Podocarpus} $\times$ \textit{Ocotea} (18.0\%), \textit{Luehea} $\times$ \textit{Ocotea} (19.2\%), and \textit{Luehea} $\times$ \textit{Nectandra} (17.4\%). In contrast, for the realization with the best average accuracy (Figure \ref{fig:hmt}), all species reached high accuracy values, with 10 species presenting accuracy of 88\% (\textit{Luehea}, \textit{Araucaria}, \textit{Mimosa}, \textit{Campomanesia}, \textit{Cedrela}, \textit{Cinnamodendron}, \textit{Cupania}, \textit{Ocotea}, \textit{Schinus sp1}, \textit{Schinus sp2}) and 4 species with values among 75\% and 84\% (\textit{Lithraea}, \textit{Matayba}, \textit{Nectandra}, \textit{Podocarpus}).

Finally, we investigated the MTFCsp model performance using  smaller training sets, selecting 14 ITCs (one ITC per species) and 23 ITCs (maximum two ITCs per species) to create the training set. The evaluation considered the same test set for all experiments. As expected, the performance is directly related to the number of training samples, notably affected for the training set with 14 ITCs (see Table \ref{tab:lessITC}). Still, considering the challenge of training the model with a single sample per species and the variability of tree crowns, the results showed to be very encouraging. Given the highly diverse forest setting in our study, those results show the potential of using the multi-task approach for training an FCN with scarce annotated ITCs samples. It is also important to emphasize that the performance achieved depends not only on the number of training samples, but also on the particularities of each study area.

\begin{table}[htbp]
\caption{MTFCsp model performance trained with different number of ITCs samples.}
\label{tab:lessITC}
\begin{adjustbox}{width=0.65\textwidth,center}
\begin{tabular}{r|ccccc}
\toprule
ITCs & avg UA (\%) & avg PA (\%) & avg F1 (\%) & Kappa (\%) & OA (\%) \\
\midrule
14 & 47.41 & 60.25 & 48.90 & 40.81 & 46.76 \\
23 & 75.42 & 79.26 & 71.93 & 68.77 & 72.55 \\
38 & 88.63 & 88.59 & 87.51 & 83.50 & 85.91 \\
\bottomrule
\end{tabular}
\end{adjustbox}
\end{table}

\subsection{Tree species map}
\begin{figure}[ht]
	\centering
		\includegraphics[width=1\linewidth]{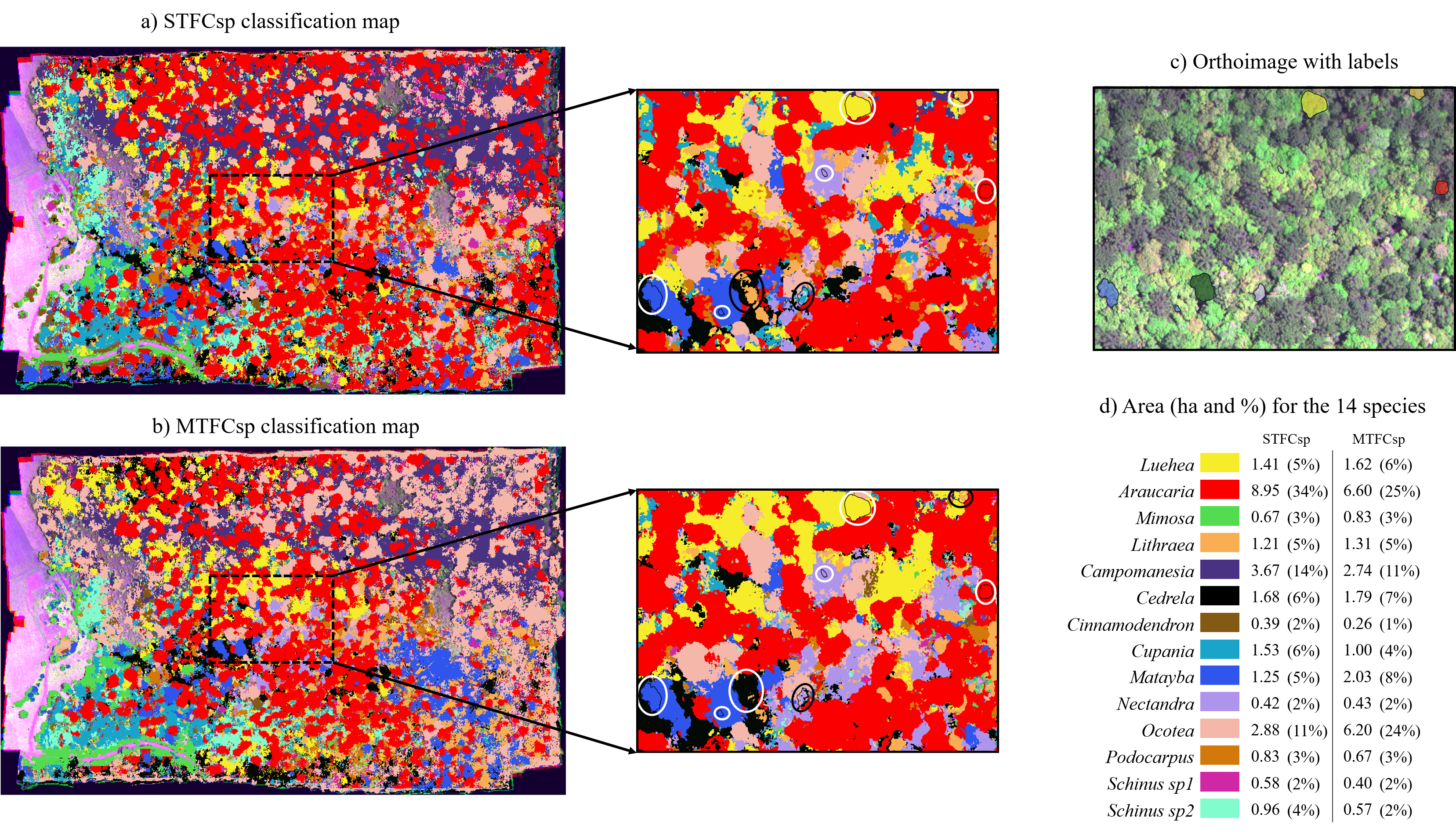}
\caption{Classification images and representative enlarged area. (a) the STFCsp classifier; (b) the MTFCsp classifier; (c) orthoimage with test ITC labels and (d) total area (ha) and \% of total area for each of the 14 species. The white and black circles show correctly classified and miss-classified ITC samples, respectively. Note: non-treed areas were removed using a canopy height model mask.}
\label{fig:visuals}
\end{figure}

Figure \ref{fig:visuals} shows the classification images for the study area. It is possible to observe that both methods produced consistent results considering the ITCs samples from fieldwork. The Figure presents a zoom in the central region of the classified images containing 8 ITC test samples for better visualization. 

Looking at the white circles in Figure \ref{fig:visuals}, it turns out that both methods correctly detected and classified most ITC samples. However, many pixels were misclassified for both methods, as shown in the black circles in Figure \ref{fig:visuals}. For the STFCsp method, we observed a \textit{Cedrela} sample with close to 50\% of pixels wrongly classified as \textit{Lithraea}, and a \textit{Nectandra} sample with most pixels classified as \textit{Matayba} and \textit{Cupania}. For the MTFCsp method, we also observed misclassification in 2 ITCs, with some pixels of \textit{Lithraea} erroneously classified as \textit{Luehea}, and \textit{Nectandra} as \textit{Cedrela} and \textit{Ocotea}, respectively. 

Besides, we noticed that MTFCsp managed to detect species but not classify all pixels accordingly every time, whereas STFCsp misclassified all the ITC pixels in some cases. A close inspection of the classification map allows to identify the abovementioned remark for \textit{Matayba} and \textit{Cedrela}, were the multi-task method correctly detected and classified these species, whilst the single-task method assigned \textit{Ocotea} and \textit{Lithraea}, respectively. As expected, in both maps, \textit{Araucaria} was the dominant species, but the MTFCsp map had more areas classified as \textit{Ocotea} than STFCsp. The representativeness of the species in the classification maps is in line with studies reporting the predominant species of the Mixed Ombrophilous Forest in Santa Catarina state. According to \citep{Higuchi2012,Manfredi2015}, this phytophysiognomy is characterized not only by the remarkable presence of \textit{Araucaria angustifolia}, which is already a common sense, but also by an important set of typical broadleaves species, such as \textit{Matayba elaeagnoides} and \textit{Ocotea pulchella}.

Finally, we observed that the MTFCsp model resulted in a more homogeneous classification map. In contrast, the STFCsp model presented more salt and pepper effect, with different species being assigned within one ITC, highlighting the benefits of the distance map to improve the classification and the segmentation of the ITCs.

\section{Discussion}\label{disc}
\subsection{Considerations about tree species classification}
This work proposed a new network architecture for semantic segmentation of tree species in dense forests from hyperspectral data using small training sets. To enable that, we proposed a partial loss function to train an FCN with scarce and sparse ITC training samples. Similar to \cite{wu2018scribble}, we found that the partial loss function improved the semantic segmentation results in our scarce label use case. Furthermore, the multi-task model introducing the distance regression branch improved the semantic segmentation accuracy and produced a visually more appealing tree species map of the study area. Numerically, we observed that the complementary task boosted the performance between 8\% and 11\% for OA, Kappa value, user's accuracy, producer's accuracy, and average F1 score. Although previous works used a regression branch in a multi-task FCN for improving the segmentation of impervious surfaces, buildings, cars, low vegetation, trees, and background from remote sensing images \citep{diakogiannis2020resunet}, also in combination with a partial loss function \citep{wang2019boundary}, we believe this is the first work that combines both techniques for tree species mapping in a dense tropical forest region. 

An essential contribution in this work is the reduction of the demand for ITC annotated samples to train an FCN that delivers dense semantic segmentation. This is a significant contribution considering the time and effort needed for identifying tree species in fieldwork. 
We further believe this proposal is also applicable to other dense forests where ITCs are typically scarce and sparsely annotated over the study area \citep{ferreira2016mapping,ferraz2016lidar,ferreira2019tree}.   

Another critical aspect regards the risk of overfitting, when annotated training sets are small such as the one used in this work. Common approaches to tackle that in tree species classification involve data splitting, k-fold cross-validation, leave-one-out cross-validation, and bootstrap-resampling \citep{Fassnacht2016}. The dataset used in our analysis consists of roughly 70 annotated ITC samples, and for some species, only one ITC was available for training and one for testing. We run 25 realizations, leaving out a small percent of the available training tiles as the validation set, and, for each realization, we applied a data augmentation procedure that allowed that training tiles from all classes, regardless of their frequency, had the same probability of composing a minibatch. Also, we used random rotations and horizontal and vertical flips online, which proved to improve the network's generalization \citep{sothe2020comparative,ferreira2020individual}. We also employed some methods for network regularization, such as drop-out and L2 norm regularization.    

The experimental protocol adopted allowed to verify that variations in the training samples due to factors, as lighting conditions and spectral changes within the tree crowns (e.g., shadows, background), significantly affected the network's performance. Note that we randomly selected 1\% hold-off of the training tiles to monitor the network and applied early stopping for each realization. As a result, the tree species were not equally classified, and some of them presented more variability in accuracy and F1 score values among the 25 realizations. Similar results have been reported in previous works, where 5-10\% of variations were reported after applying iterative data-sampling \citep{Fassnacht2016}. Our experiments also revealed that species with high variability in the crown's size have the classification performance affected in the multi-task approach. Since the distance map estimation considers the object size, this introduces a new source of variability. Nonetheless, overall, the introduction of the complementary task significantly improved the classification map.

\subsection{Considerations about the distance map estimation as regularizer}
Regarding the role of the complementary task as a regularizer, we observed that the distance map estimation introduces an inductive bias into the learning process \citep{ruder2017overview}, which leads the model to learn features capable of explain both tasks and therefore generalize better. This inductive bias acts as a regularization term reducing the risk of overfitting for the main task. While both networks fit the training ITC samples almost perfectly, generally, MTFCsp performs better on the test ITC samples, which points to a model with better generalization \citep{zhang2018deep}. 

\begin{figure}
\begin{subfigure}{.5\textwidth}
  \centering
  \caption{STFCsp probability map}
  \includegraphics[width=.98\linewidth]{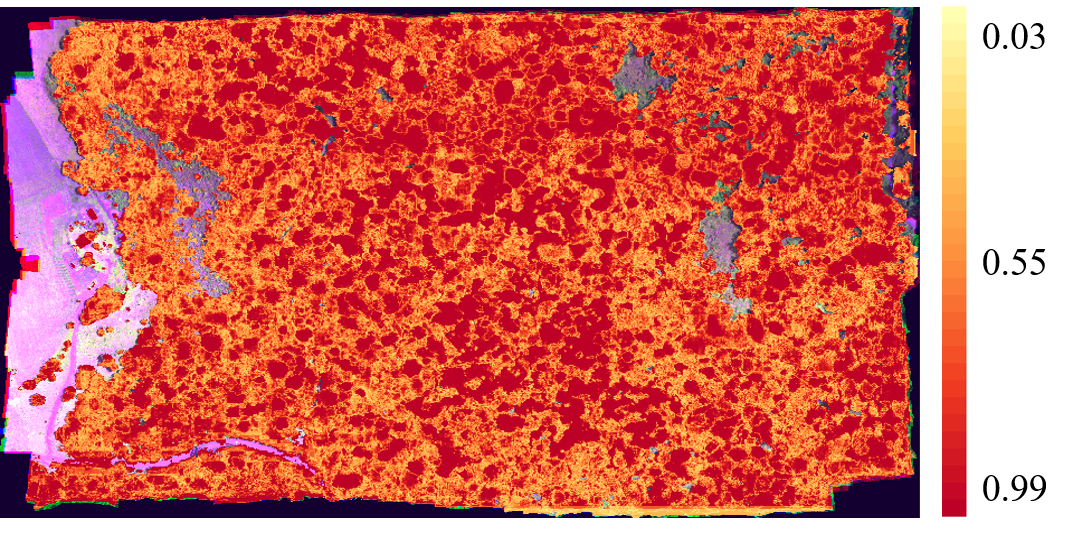}
  \label{fig:stprob}
\end{subfigure}%
\begin{subfigure}{.5\textwidth}
  \centering
  \caption{MTFCsp probability map}
  \includegraphics[width=.98\linewidth]{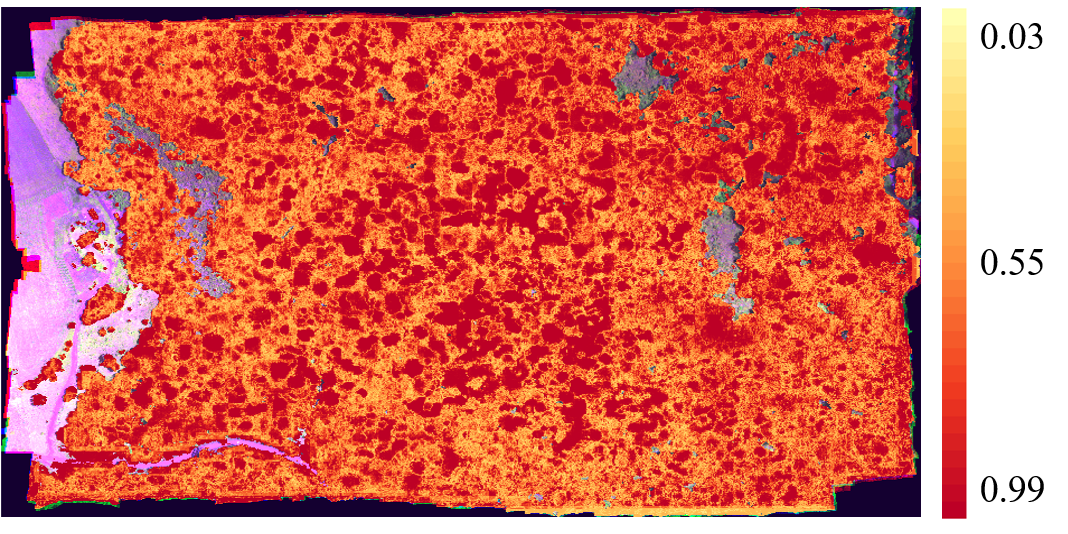}
  \label{fig:mtprob}
\end{subfigure}
\caption{Probability output for the realization with the best average class accuracy (AA) for both methods. Note: non-treed areas were removed using a canopy height model mask.}
\label{fig:entropy}
\end{figure}

\begin{figure}
\begin{subfigure}{.19\textwidth}
  \centering
  \caption{\textit{Luehea}}
  \includegraphics[width=.98\linewidth]{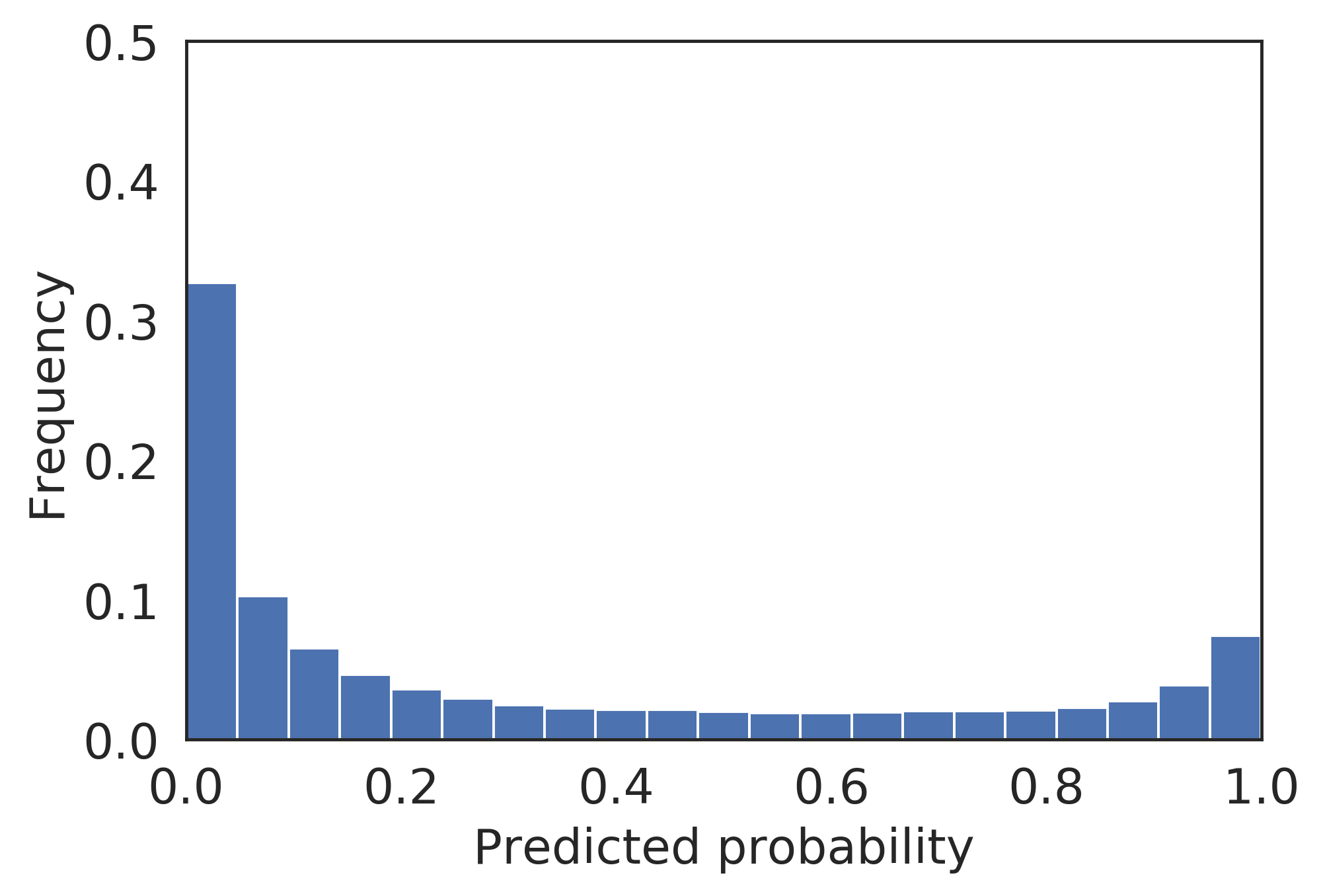}
\end{subfigure}%
\begin{subfigure}{.19\textwidth}
  \centering
  \caption{\textit{Mimosa}}
  \includegraphics[width=.98\linewidth]{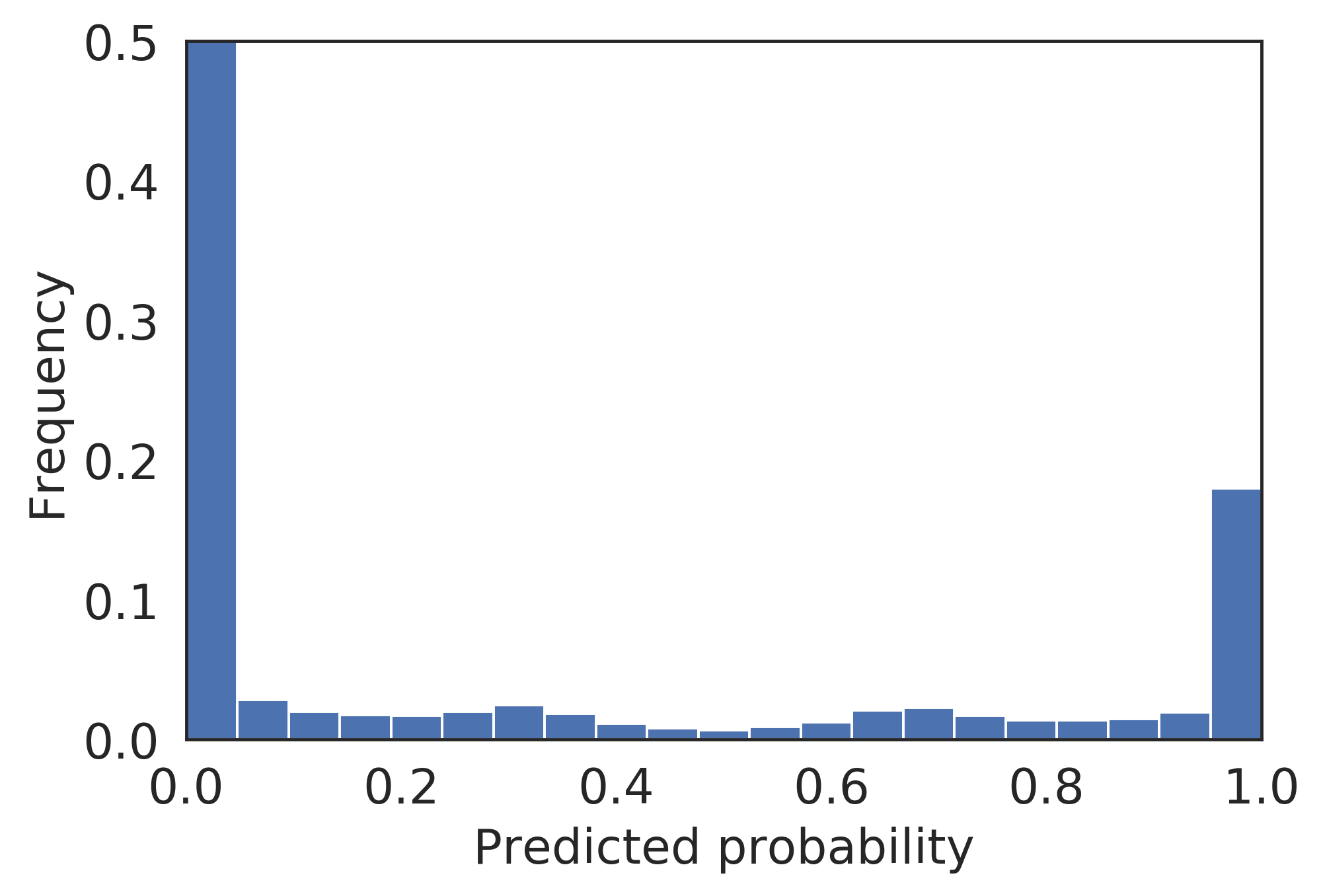}
\end{subfigure}%
\begin{subfigure}{.19\textwidth}
  \centering
  \caption{\textit{Cupania}}
  \includegraphics[width=.98\linewidth]{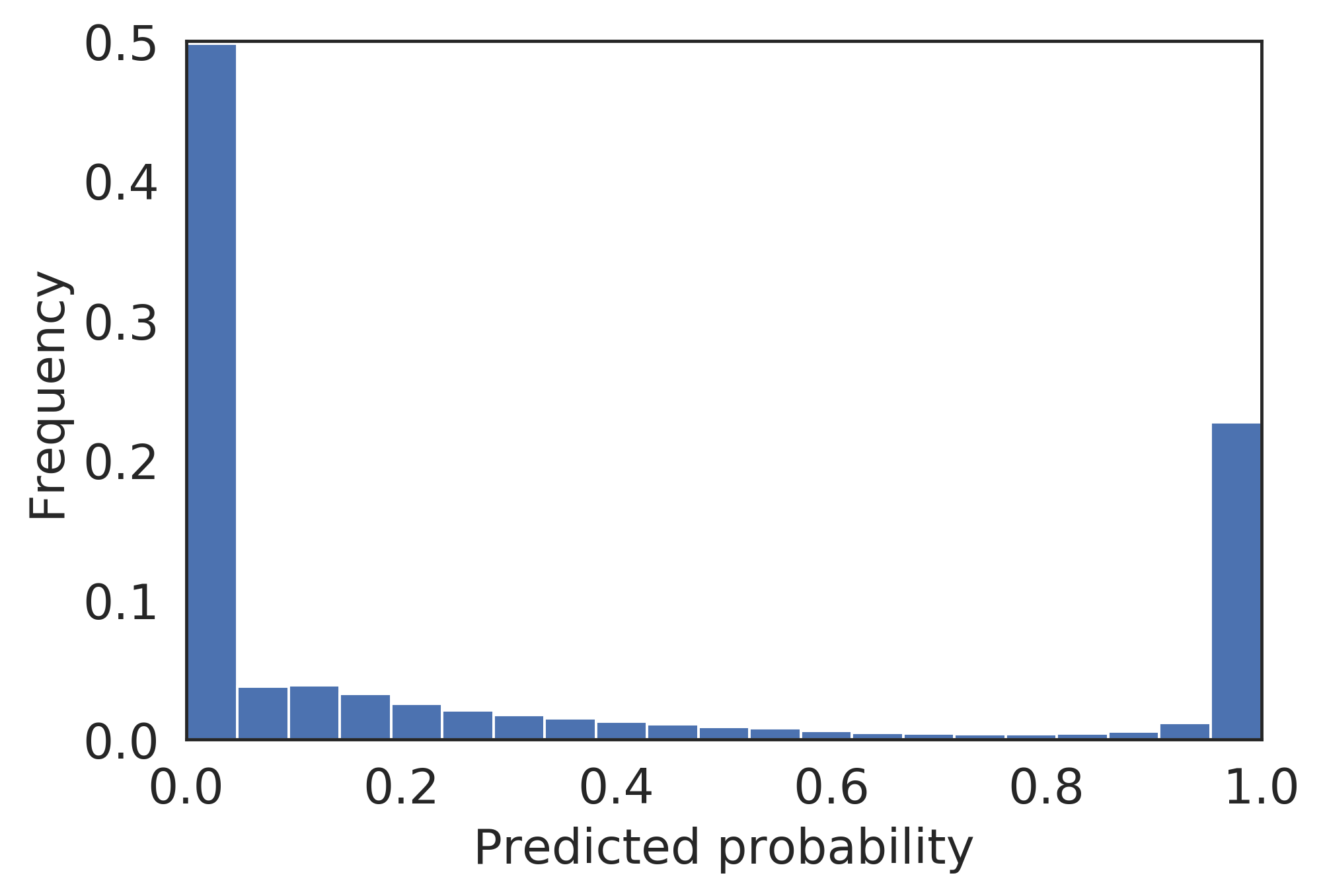}
\end{subfigure}%
\begin{subfigure}{.19\textwidth}
  \centering
  \caption{\textit{Matayba}}
  \includegraphics[width=.98\linewidth]{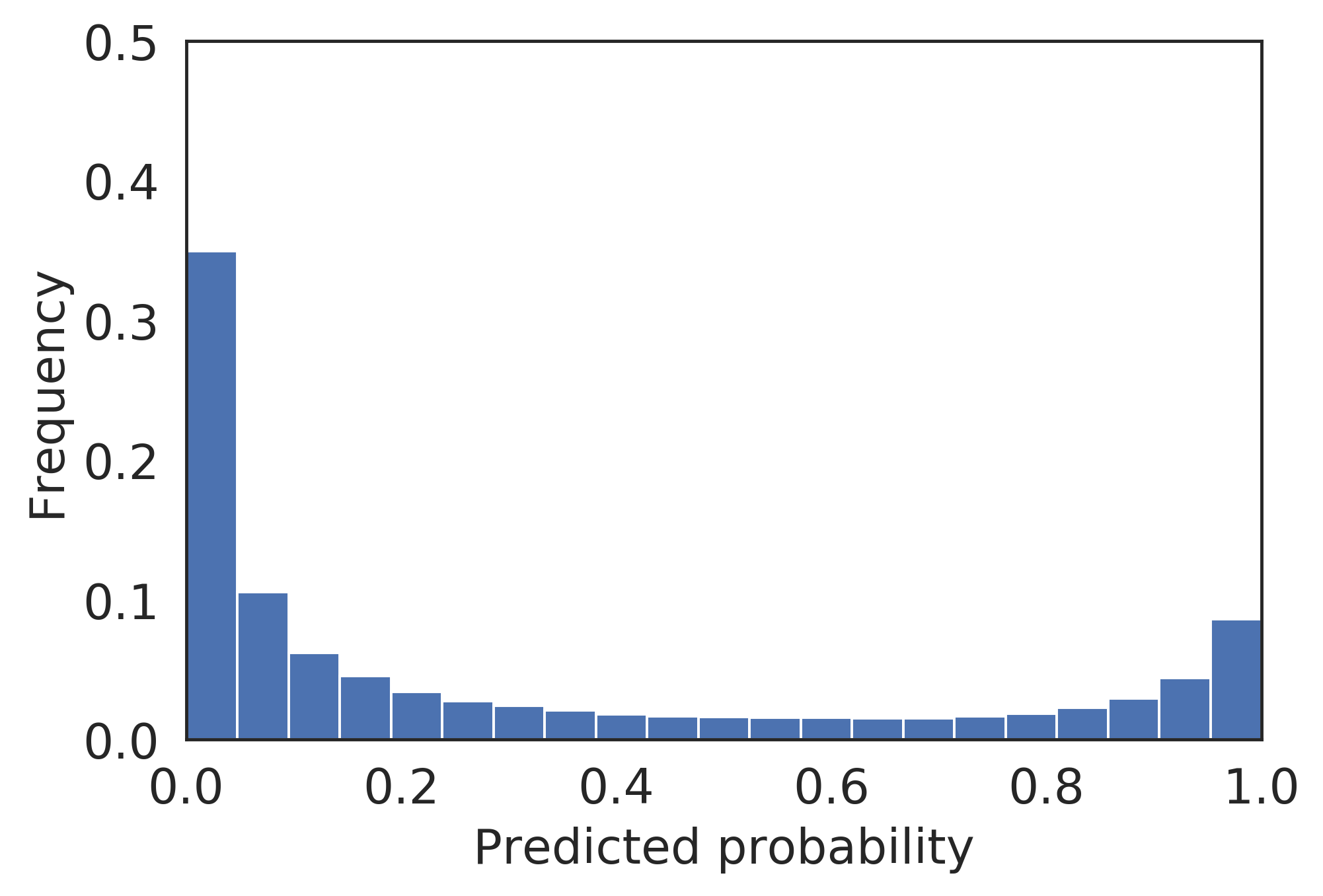}
\end{subfigure}%
\begin{subfigure}{.19\textwidth}
  \centering
  \caption{\textit{Schinus sp2}}
  \includegraphics[width=.98\linewidth]{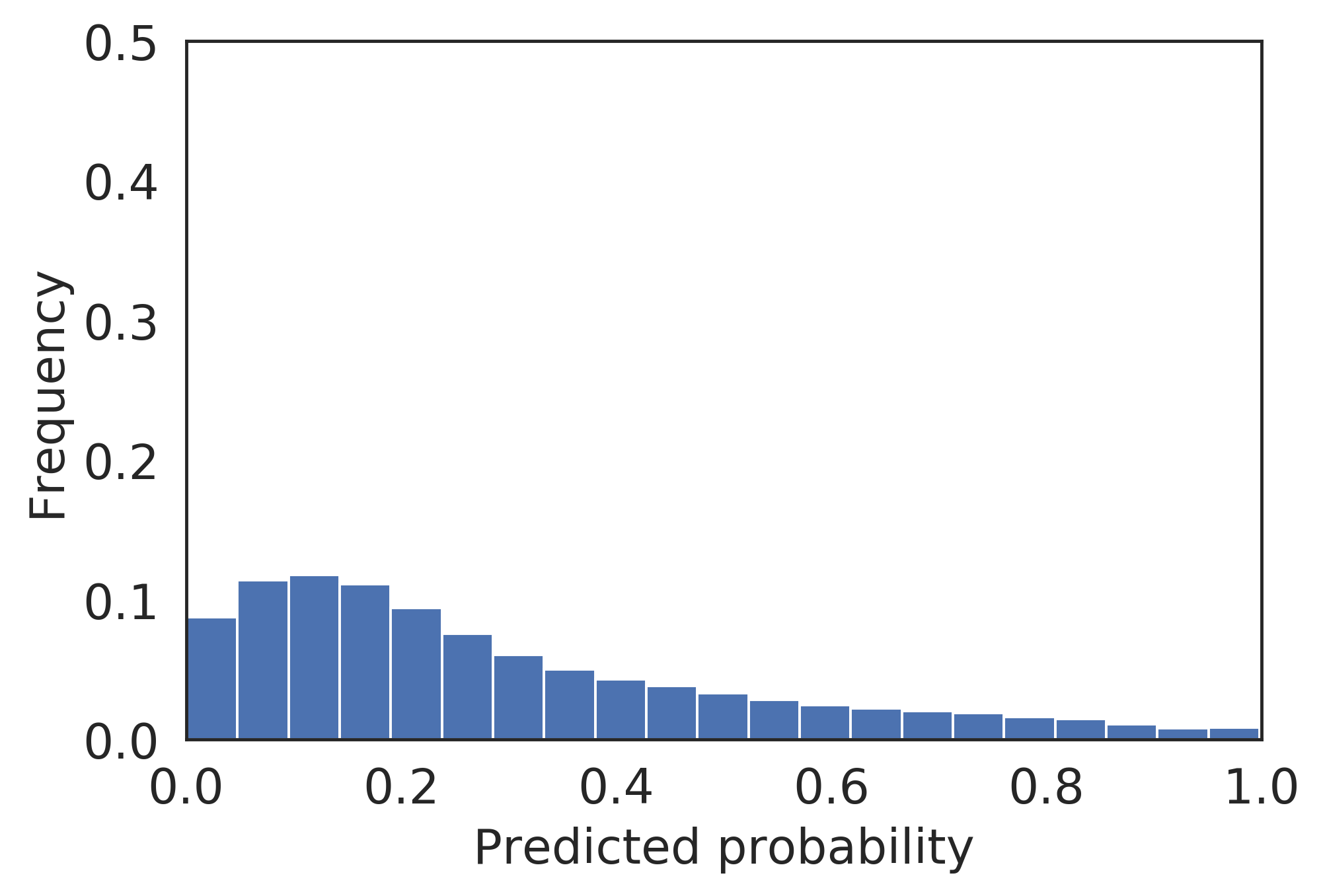}
\end{subfigure}\\
\vspace{0.5cm}
\begin{subfigure}{.19\textwidth}
  \centering
  \includegraphics[width=.98\linewidth]{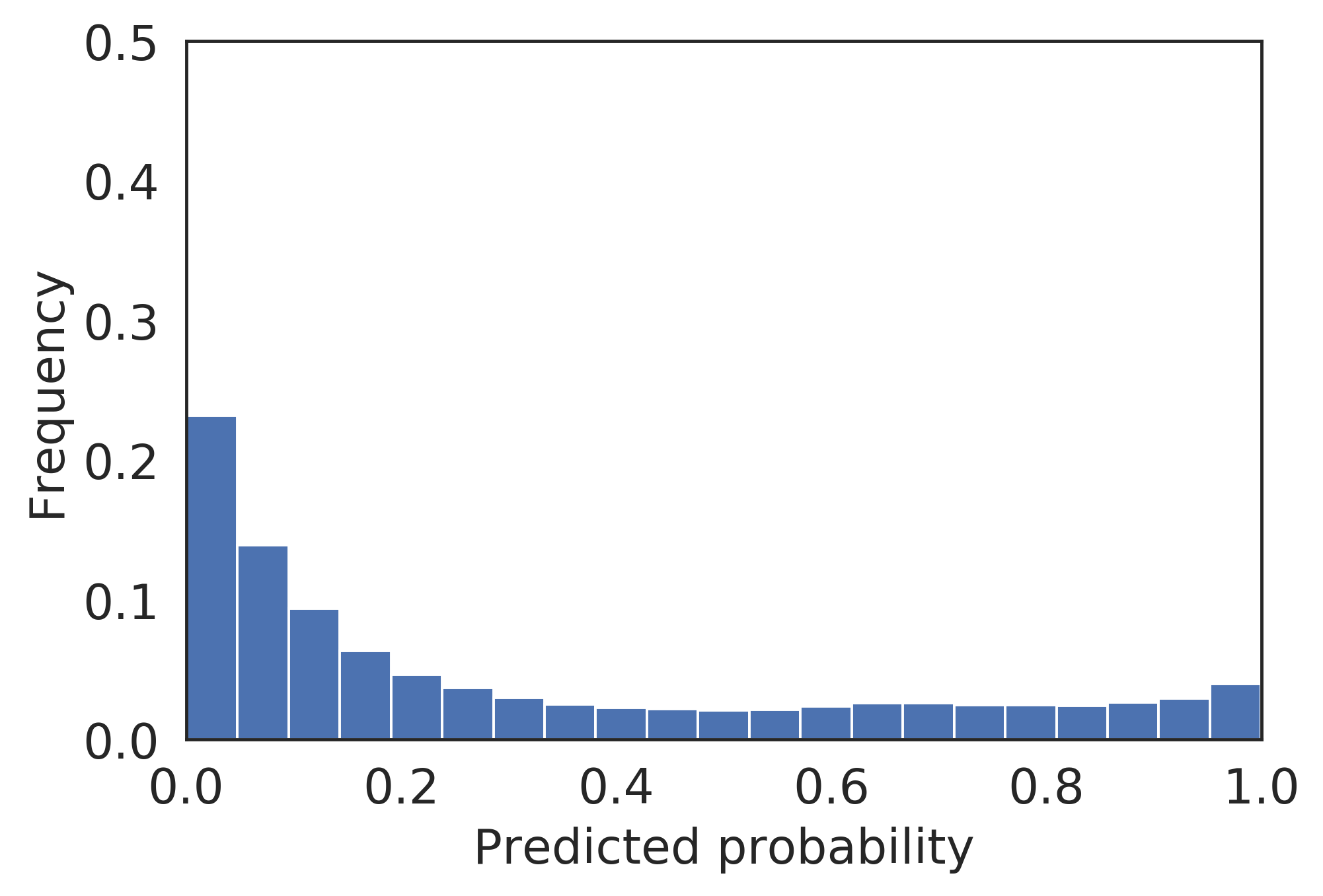}
\end{subfigure}%
\begin{subfigure}{.19\textwidth}
  \centering
  \includegraphics[width=.98\linewidth]{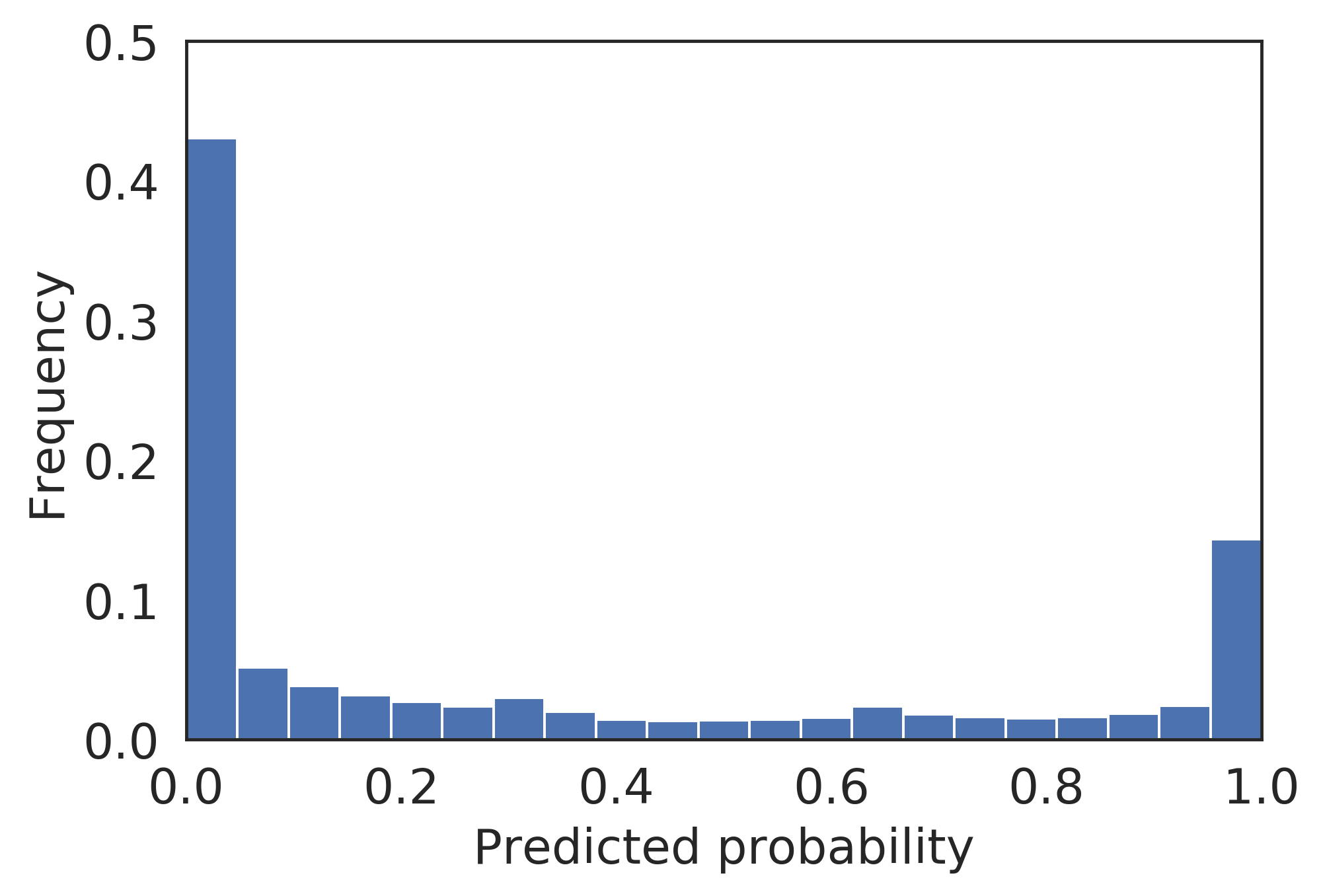}
\end{subfigure}%
\begin{subfigure}{.19\textwidth}
  \centering
  \includegraphics[width=.98\linewidth]{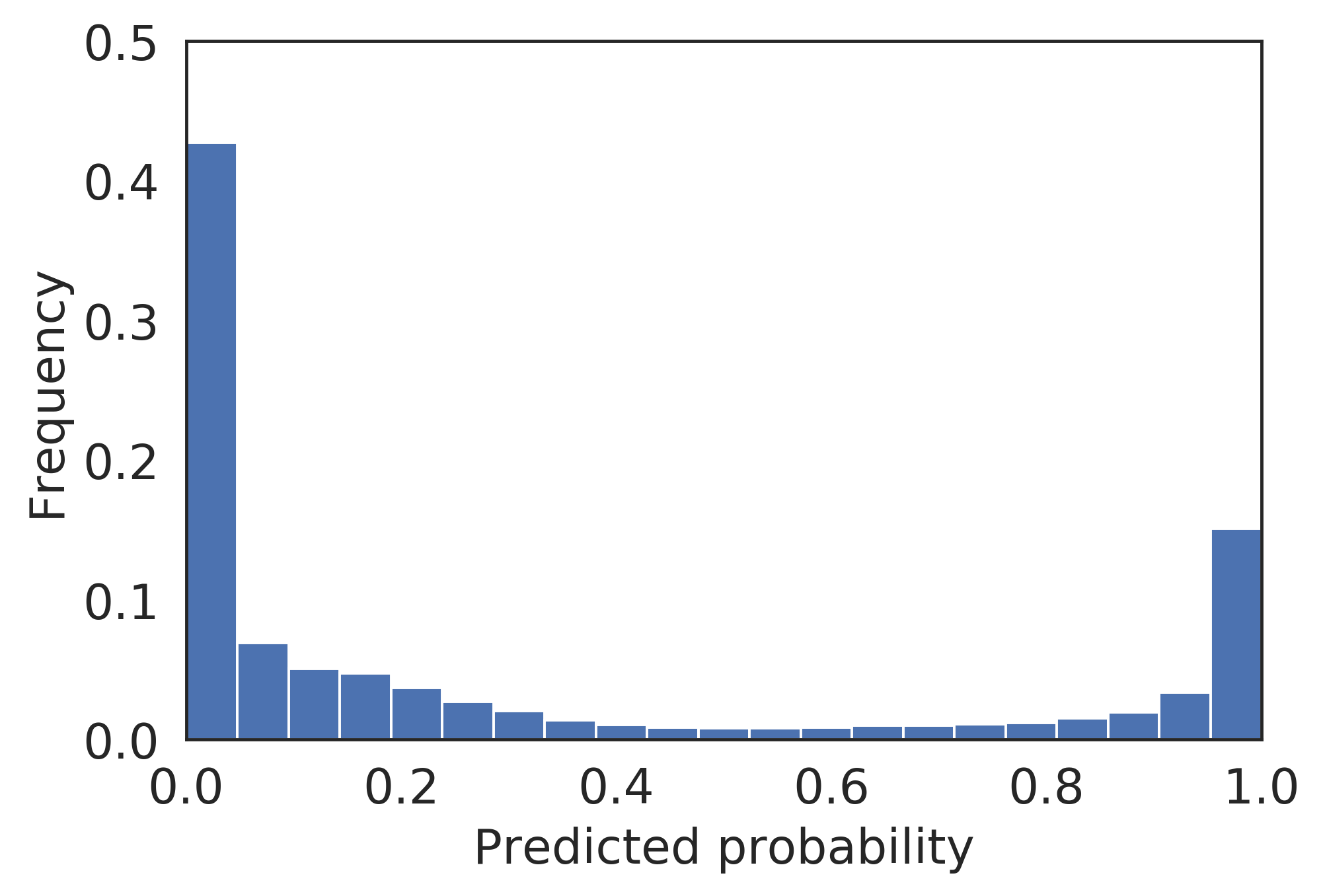}
\end{subfigure}%
\begin{subfigure}{.19\textwidth}
  \centering
  \includegraphics[width=.98\linewidth]{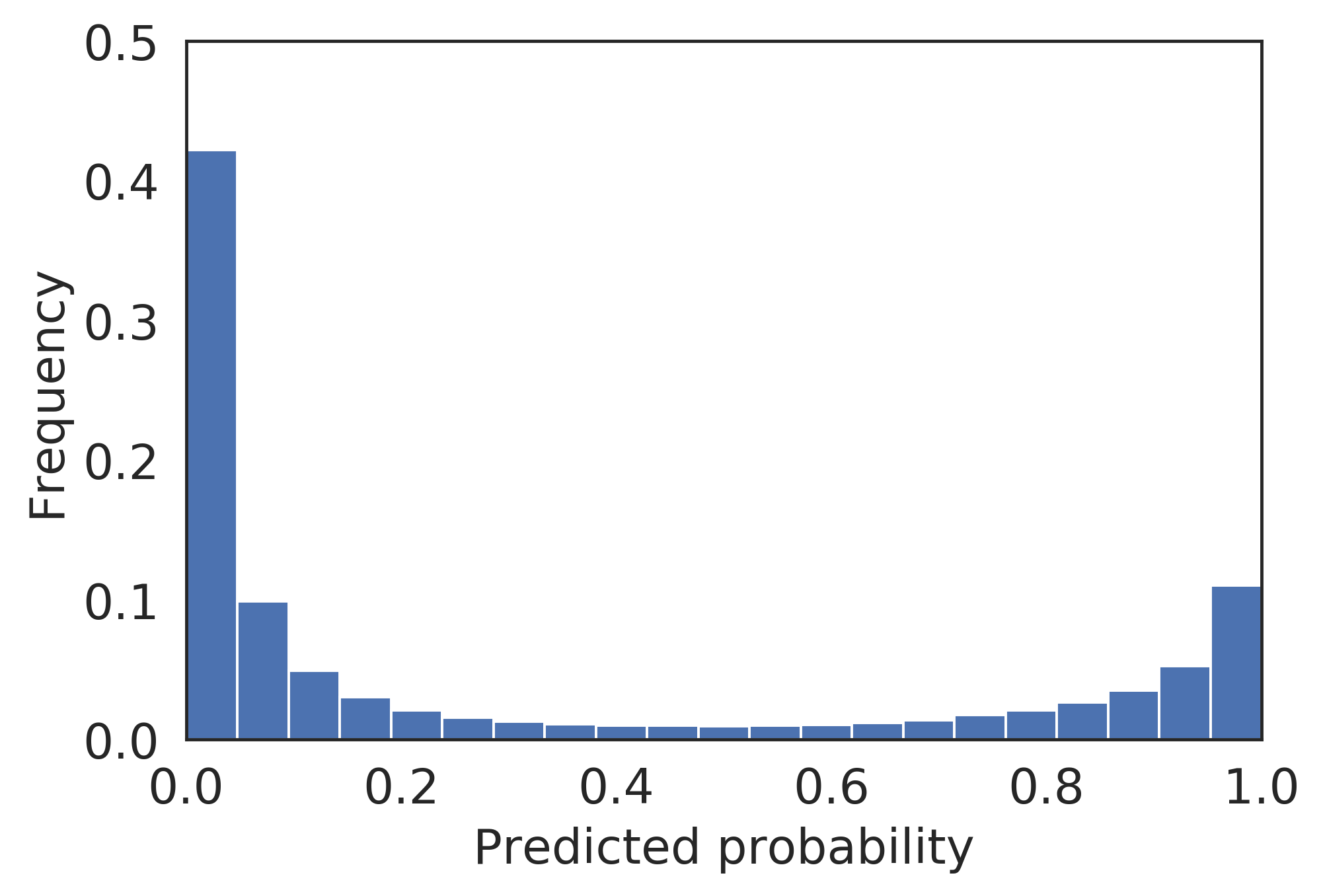}
\end{subfigure}%
\begin{subfigure}{.19\textwidth}
  \centering
  \includegraphics[width=.98\linewidth]{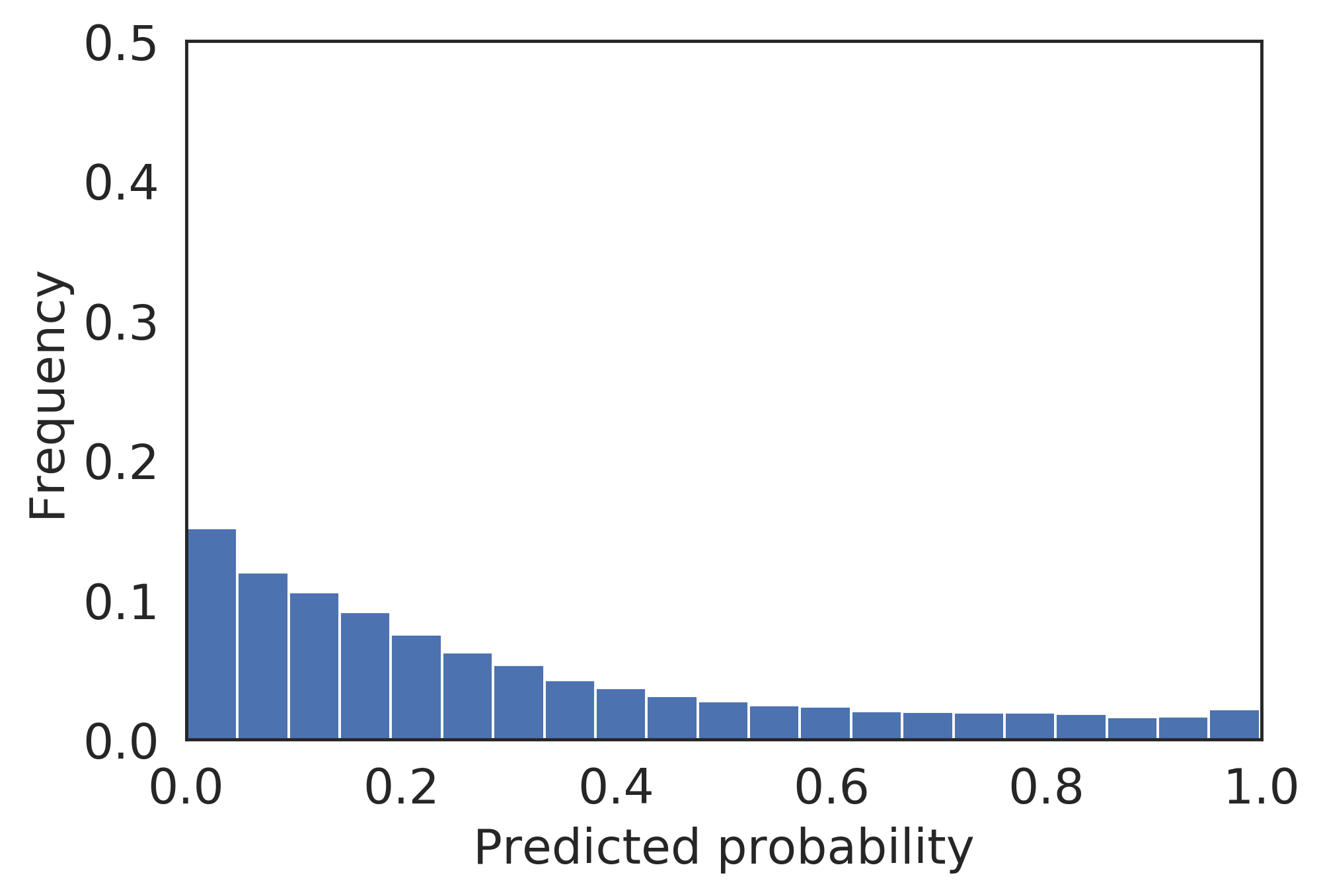}
\end{subfigure}
\caption{Distribution of the magnitude of softmax probabilities on the correctly classified test pixels for some of the species for both methods. Top row: STFCsp, bottom row: MTFCsp method.}
\label{fig:entropclass}
\end{figure}

Figure \ref{fig:entropy} shows the probability output for each method's predicted class, and one can note that the single-task model delivers results closer to one-hot predictions, especially if compared to MTFCsp, and such low entropy outcomes are often regarded as an indicator of overfitting \citep{szegedy2016rethinking}. To better understand the networks' behavior, we also report the histogram of softmax probabilities (considering the top 3 classes ranked) for some of the species (see Figure \ref{fig:entropclass}). Generally, the STFCsp model leads to softmax distributions with a higher concentration of 0 and 1 probability values. MTFCsp model, on the other hand, delivers smoother output distributions (see \textit{Luehea}, \textit{Mimosa} and \textit{Cupania} in Figure \ref{fig:entropclass}). Even if that was observed for 8 species, 6 species presented the opposite behavior, including \textit{Matayba} and \textit{Schinus sp2} (see Figure \ref{fig:entropclass}). Overall, the global entropy of the MTFCsp method was 8\% higher than the STFCsp method, which indicates that the complementary task can also act as entropy-regularization for the architecture.

\subsection{Potential use of the distance map for ITC detection and delineation}
Analyzing and processing the distance map prediction is beyond the scope of this paper. However, we can consider its potential use for other tasks, such as ITC detection, tree count, ITC delineation, ITC statistics, and others. Information about tree counts in tropical forests play an essential role in understanding forest diversity and dynamics \citep{davies2021forestgeo,cavaleri2015urgent}. Many applications require an accurate estimation of the total number of trees, including protection of natural forests, wildlife habitat mapping, conservation, and forest management \citep{onishi2021explainable}. However, estimating the tree count from a classification map is challenging, particularly in forest regions where tree crowns often overlap. As observed in Figure \ref{fig:visuals}, the classification map fails to identify individual trees when neighboring trees pertain to the same species. In contrast, the distance map predicted by the second branch of the multi-task method could be potentially used to count the number of trees using their centers, as highlighted in yellow.

ITC delineation is a prerequisite for individual tree inventory over large spatial extents \citep{clark2005hyperspectral}, providing information such as tree location, crown size and distance between individuals \citep{Fassnacht2016}. Nevertheless, most of the studies exploring ITC delineation have been developed for temperate or boreal forests \citep{ke2011comparison,duncanson2014efficient,dalponte2015delineation,lee2017graph}, and their application to deciduous and tropical forests has proven to be much more challenging \citep{tochon2015use,wagner2018individual}. We advocate that the distance map estimated in our approach may be further explored for ITC delineation in highly diverse forests. 

\begin{figure}
	\centering
		\includegraphics[width=0.98\linewidth]{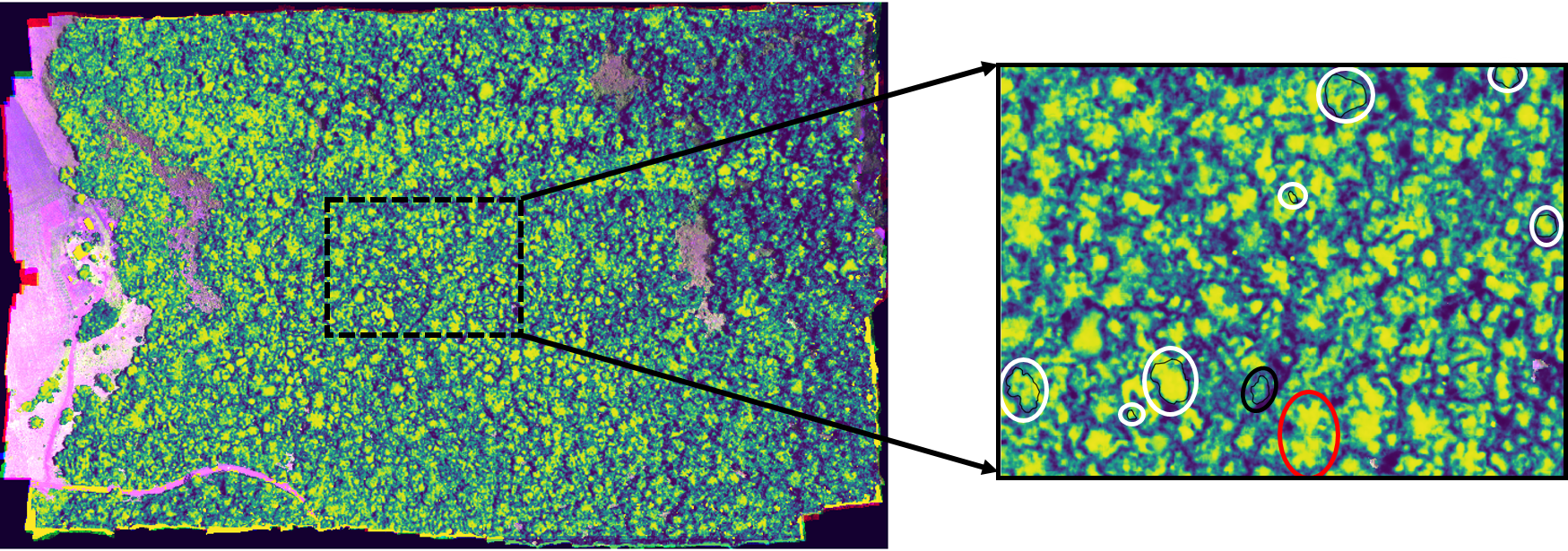}
	\caption{Distance map resulted from the regression branch of the proposed MTFCsp method and representative enlarged area with the overlaid contour of 8 test ITC samples. Yellow pixels are located in the crowns’ center region, while dark blue pixels are located at the crowns’ edges. White circles indicate correctly delineated ITCs, while the black circle indicates a bad delineation result. The red circle indicates two ITCs of the species \textit{Araucaria} that were correctly delineated}
	\label{fig:depth}
\end{figure}

In Figure \ref{fig:depth} we report the distance map where the white circles indicate 7 test ITCs correctly detected and delineated, and the black circles correspond to ITC samples with lower detection values. A visual inspection of the prediction map reveals that the great majority of the trees were correctly detected (yellow pixels located in the crowns’ center region), and the crowns’ edges and background regions were also identified (dark blue pixels). Another important finding is that the distance map roughly delineates the shape of a species characterized by a specific crown geometry, e.g., \textit{Araucaria} (red circle in Figure \ref{fig:depth}). That species, for instance, is critically endangered according to the “List of Threatened Species” of the International Union for Conservation of Nature (\textbf{IUCN 2017}) \citep{iucn2017guidelines} and, therefore, knowing the number of those trees and their spatial distribution would be very attractive for conservation purposes. 

\subsection{Considerations about the proposed method and baseline counterparts performances}

The proposed method, MTFCsp, outperformed the baseline CNN approach and traditional machine learning approaches using RF and SVM. As pointed by \citet{li2017spectral} and \citet{gao2018hyperspectral}, the main advantage of deep learning methods, such as CNN, consists in their ability to extract spatial and spectral features automatically from the original images and learn features through training, with minimal prior knowledge about the task.

The study conducted by \citet{sothe2020comparative}, reported that the RF and SVM only reached similar accuracies to those observed in CNN patch-based method when incorporating 3D information extracted from the UAV-photogrammetric point cloud and using a final classification that aggregated the classification inside segments using a majority vote rule. That suggests neighboring context plays a vital role in the final pixel-wise classification. Therefore, we conjecture that deep learning methods reach higher accuracies, primarily due to the contextual window (patch sizes) used for feature classification, which considers the information on the pixel neighborhood. In our study case, the spatial structure of canopies is related to the tree size: if a pixel falls in a specific tree, its neighbors are also likely to be in the same tree and have similar information. Deep learning classifiers operate according to such principle of detecting patterns in groups of adjacent pixels and relating them to background information \citep{fricker2019convolutional}.

An essential advantage of FCN approaches compared to CNN patch-based approaches relates to computational cost at inference time. CNN patch-based methods often need to decompose input images into a series of overlapping tiles to further predict the class for the central pixel of the patch, implying high computational cost and redundant operations for adjacent pixels. Moreover, patch-based approaches usually assume predictions are spatially independent \citep{volpi2016dense}, which impacts the performance when contextual dependencies in space are present. In contrast, FCN models, such as those proposed in this paper use a single forward pass to predict every pixel of an input tile. Furthermore, FCN prediction considers learned spatial dependencies between neighboring pixels at inference. As discussed in \citet{volpi2016dense}, predicting the label for all pixels within a tile in a single forward step is remarkably efficient, which is vital for large-scale remote sensing image analysis. As shown in Table \ref{tab:time}, the CNN approach takes more than two hours to classify the whole study area (6570$\times$4043 pixels) using a GPU, whereas the STFCsp and MTFCsp methods takes 158 s and 174 s, respectively, considering the three overlapping ratios studied.

\begin{table}[htbp]
\caption{Inference time to process the whole study area ($\approx$ 26,5 millions pixels).}
\label{tab:time}
\begin{adjustbox}{width=0.65\textwidth,center}
\begin{tabular}{r|c|ccc|c}
\toprule
 & \multirow{2}{*}{CNN} & \multicolumn{4}{c}{STFCsp/MTFCsp} \\
 & & 10\% & 30\% & 50\% & Total \\
\midrule
time in s & 8,300 & 35/38 & 46/51 & 77/85 & 158/174 \\
tiles processed & 26,5 millions & 1,995 & 3,404 & 6,656 & 12,055\\
\bottomrule
\end{tabular}
\end{adjustbox}
\end{table}

\section{Conclusion}\label{conclusion}
This paper presented a novel methodology to train FCNs from a sparse and scarce ITC sample set for tree species classification in dense forests canopies such as that found in tropical regions. We proposed a partial loss function to train a multi-task network that uses an auxiliary task for distance map regression to improve the semantic segmentation performance. Our results demonstrated that including the complementary task improved the semantic segmentation performance by more than 11\% in user's accuracy and more than 7\% in producer's accuracy compared to the single-task counterpart.

Our proposed deep learning architecture uses an encoder composed of residual blocks units and two decoders, one for each task. The semantic segmentation decoder builds on the DeepLabv3+ architecture with parallel atrous convolutions and image pooling, whereas the distance map decoder builds on traditional convolution operations. We introduced a partial loss function for training the models, which only back-propagates the losses from pixels that pertain to the annotated region and enables training dense semantics segmentation networks from weakly annotated samples. 

We experimented two FCN approaches using UAV-hyperspectral data for tree species classification in a subtropical forest area. The first one, a single-task network (STFCsp), outperformed the RF and SVM baseline methods without the need of relying on hand-engineered features, reaching an overall accuracy (OA) of 77.29\%. However, the performance was lower than the one observed in a CNN baseline method (OA of 83\%), highlighting the burden of training FCN with sparse and scarce ITC samples. Nonetheless, the STFCsp reached similar average user's and producer's accuracy to CNN approach with considerably lower processing time and computational costs. In the second approach, a multi-task model (MTFCsp), we included a complementary task that led to considerably gains in accuracy (OA of 85.91\%), outperforming not only the single-task counterpart but also the CNN baseline. Being a FCN architecture, the proposed approach also demands less computational effort at inference time than CNN patch-based approaches. 

We also demonstrated that the distance map transform acts as a regularizer, improving the semantic segmentation generalization. Moreover, we discussed how the distance map predictions could help ITC detection and delineation and tree species classification, which ultimately helps monitoring forest biodiversity and endangered tree species, like \textit{Araucaria} and \textit{Cedrela}, or detecting invasive species. 

Finally, we point to further studies regarding model generalization since the network architecture design may have to be adjusted to different forest types. Testing the method in other diverse areas will enable more conclusive and generic outcomes.

\section*{Acknowledgments}

We gratefully acknowledge the financial support offered to this research by the Brazilian National Council for Scientific and Technological Development (CNPq); the Foundation for Support of Research and Innovation, Santa Catarina State (FAPESC), the São Paulo Research Foundation (FAPESP) and the Foundation for Support of Research and Innovation, Rio de Janeiro State (FAPERJ). 




\clearpage
\bibliography{biblio.bib}

\end{document}